\documentclass[12pt, draftclsnofoot, onecolumn]{IEEEtran}

\usepackage{cite}
\usepackage{algorithmic}
\usepackage{array}
\usepackage{bbm}
\usepackage{amsfonts}
\usepackage{graphicx}
\usepackage{optidef}
\usepackage[hidelinks]{hyperref}
\usepackage{xcolor}
\hypersetup{
	colorlinks  = true,
	linkcolor	={red!100!black},
	citecolor	={blue!100!black},
	urlcolor	={blue!100!black}
}
\usepackage{mathtools}
\usepackage{subfigure,epsfig}
\usepackage{amsmath, amsthm, amssymb}
\usepackage[outdir=./]{epstopdf}
\usepackage{dsfont}
\usepackage{mathrsfs}
\usepackage[ruled,vlined]{algorithm2e}
\SetKwInput{kwInit}{Initialization}
\newtheorem{Theorem}{Theorem}
\newtheorem{Corollary}{Corollary}

\newtheorem{Lemma}{Lemma}
\newtheorem{Assumption}{Assumption}

\def\figref#1{Fig.\,\ref{#1}}%
\newlength{\figwidth}
\begin{document}
\title{Federated Learning in Unreliable and Resource-Constrained  Cellular Wireless Networks}
\author{Mohammad Salehi and Ekram Hossain, \textit{FIEEE}\thanks{The authors are with the Department of Electrical and Computer Engineering at the University of Manitoba, Canada 
		(emails: salehim@myumanitoba.ca, Ekram.Hossain@umanitoba.ca). \textbf{E. Hossain} is the corresponding author. The work was supported by a Discovery Grant from the Natural Sciences and Engineering Research Council of Canada (NSERC).
}}

\maketitle

\begin{abstract} With growth in the number of smart devices and advancements in their hardware, in recent years, data-driven machine learning techniques have drawn significant attention. However, due to privacy and communication issues, it is not possible to collect this data at a centralized location. Federated learning is a machine learning setting where the centralized location trains a learning model over remote devices. Federated learning algorithms cannot be employed in the real world scenarios unless they consider unreliable and resource-constrained nature of the wireless medium. In this paper, we propose a federated learning algorithm that is suitable for cellular wireless networks. We prove its convergence, and provide the optimal scheduling policy that maximizes the convergence rate. We also study the effect of local computation steps and communication steps on the convergence of the proposed algorithm. We prove, in practice, federated learning algorithms may solve a different problem than the one that they have been employed for if  the unreliability of wireless channels is neglected. Finally, through numerous experiments on real and synthetic datasets, we demonstrate the convergence of our proposed algorithm.  
\end{abstract}


\section{Introduction}
\subsection{Motivation}
With the rapid growth in Internet-of-Things (IoT) applications and increase in the computational and storage power of smart devices, modern distributed networks generate a huge amount of data everyday \cite{li2020magazine}. Owing to this reason, data-driven machine learning techniques have gained significant attention in recent years. Currently, most of the existing machine learning techniques are centralized, i.e. they assume all data is available at a centralized location, where a central processor trains a powerful learning method on the data \cite{wang2019adaptive}. However, transferring data from user devices to the centralized location violates users' privacy \cite{khan2020federated}. To cope with this issue, federated learning has been introduced where a learning model is trained over remote devices under the control of the centralized location, called server \cite{li2020magazine, khan2020federated,kairouz2019advances}. Specifically, in federated learning (FL), the server broadcasts the global model parameters to the remote devices. Each remote device uses its local dataset to update the global model, and then transmits the updated local model to the server. After aggregating the local models, the server updates the global model and repeats the whole procedure. As an example, consider the task of next-word prediction on mobile phones, where a language model predicts the most probable next word or phrase based on a small amount of user-generated preceding text. To maintain the users privacy, instead of transmitting the raw text data to the server and training a predictor at the server, we use federated learning \cite{hard2018federated,li2020magazine}.

Although the above definition of federated learning seems similar to parallel optimization and distributed machine learning in datacenters, due to the following challenges, it needs to be treated separately: i) In federated learning, connections between the remote devices and the server\footnote{In a cellular wireless network, the base stations (BSs) can act as the servers.} are unreliable and slow, ii) different devices in the network have different systems characteristics (systems heterogeneity), and  iii) training data are not independently and identically distributed (statistical heterogeneity) \cite{li2019convergence,karimireddy2020scaffold}. Since data is non-i.i.d. (not independently and identically distributed) across devices, all devices must participate in the learning process. However, due to the systems heterogeneity, which includes variable computation and communication capabilities at different devices, and limited amount of available resources such as bandwidth (e.g. number of resource blocks)  for communication, full device participation at each round of communication is not possible. Moreover, in reality, transmission success probability is different for different devices, even when they all have the same hardware. Specifically, transmission success probability for devices that are located closer to the server is generally higher than that for devices that are located far. Thus, unless  this issue is considered at the time of updating the global model, the updated global model will be biased towards cell centre devices' local models.  

\subsection{Related Works}
To reduce the communication rounds of FL, three main approaches have been employed: quantization, sparsification, and local updates \cite{amiri2020machine}. In this paper, we study local updates, where between any two aggregation steps in  consecutive rounds, each device performs multiple local update steps. In this regard, \cite{wang2019adaptive, dinh2019federated, zhou2017convergence}  studied the convergence of communication-efficient FL with local updates for convex and non-convex problems. However, it was assumed that all of the devices participate in the aggregation step, which is obviously  not possible when the number of available resource blocks is limited. 
To tackle this problem, in \cite{mcmahan2017communication, li2020federated, li2019convergence, karimireddy2020scaffold,haddadpour2019convergence,nguyen2020fast}, at the beginning of each round, the server samples a subset of devices and allocates the available resource blocks to these devices. After performing local update steps, the BS aggregates the local models of the chosen (scheduled) devices and updates the global model. The works in \cite{mcmahan2017communication, li2020federated, li2019convergence, karimireddy2020scaffold,haddadpour2019convergence,nguyen2020fast} assumed that the BS successfully receives the local models of all the scheduled devices, and they designed the global model update step based on this assumption. However, this is not true in reality. In fact, not only is the success probability less than one, but also it is different for different devices. Clearly, to update the global model we need to include both scheduling policy and success probability. 

To better understand, consider a scenario with two devices and two resource blocks, where both devices are scheduled for sharing their updated local models with the server. Assume that the success probability is 1 for {\em device one} and is 0.1 for {\em device two}. In this scenario, the server receives the local model of {\em device two} once in every ten rounds on average, while the local model of {\em device one} is always successfully received. Obviously, without considering this aspect of wireless communications, the global model is biased towards {\em device one} at the end of the learning process.

To the best of our knowledge, only the work in \cite{yang2019scheduling} considered the effect of success probability on the convergence  of FL. Similar to \cite{smith2017cocoa}, \cite{yang2019scheduling} solved the FL problem using primal-dual optimization method. However, when strong duality is not guaranteed, this method may not be useful \cite{li2020federated}. Also, in the analysis, the FL global objective is assumed to be a function of linear combination of model parameters and the input features. Thus, convergence analysis for this method cannot be extended to other machine learning techniques. 

\subsection{Contributions and Organization}
The most common federated learning algorithm in the literature is FedAvg (federated averaging) \cite{mcmahan2017communication}, which lacks convergence analysis. Several works in the literature made steps towards analyzing convergence of FedAvg by modifying the algorithm. For example, \cite{li2019convergence} proposed a different averaging scheme for FedAvg, and derived the convergence rate of the modified algorithm. \cite{li2020federated} proposed FedProx, which is a generalization of FedAvg obtained by adding a proximal term to the local objectives. Recently, \cite{karimireddy2020scaffold} has proposed SCAFFOLD which improves FedAvg by adding a correction term, called client drift, to local updates. However, as we discussed in the previous subsection, these works and all other related works do not consider the effect of transmission success probability. 

In the above context, the major contributions of this paper can be summarized as follows:
\begin{itemize}
	\item We propose an FL algorithm that is suitable for unreliable and resource-constrained wireless systems. In particular, for an FL system in a cellular wireless network, the BS selects a subset of devices at each round and updates the global model based on their updated local models and success probabilities. We use stochastic geometry tools to approximately calculate the success probability for each device.
	\item Our proposed FL algorithm solves the FL problem in the primal domain. We prove that, for strongly convex and smooth problems, the algorithm converges on non-i.i.d data with rate $\mathcal{O}(\frac{1}{T})$.
	\item We study two difference scheduling (sampling) policies. We also explain how to achieve the optimal scheduling policy based on the derived convergence rate.
	\item We study the effect of number of computation steps and communication steps on the convergence rate.
	\item We show that the existing works, which do not include the transmission success probability in the global model update step (e.g. those in \cite{mcmahan2017communication, li2020federated, li2019convergence, karimireddy2020scaffold}), will not be suitable for a wireless communication environment, since they may converge to the solution of a different FL problem when the success probability is different for different devices.
	\item We verify the convergence of our proposed FL algorithm by experimenting over real and synthetic datasets. We also compare our results with centralized full batch gradient descent which can be considered as a benchmark for our algorithm.
\end{itemize}

The organization of the rest of the paper is as follows. In Section II, we introduce the system model and propose our FL algorithm. Then, in Section III, we provide the convergence analysis of our FL algorithm. Further analysis of the proposed FL algorithm and  comparison with related algorithms are provided in Section {IV}. In Section V, we present the simulation results. Finally, Section VI concludes the paper. A summary of the major notations used in the paper is given in Table~\ref{Table1}.

\begin{table}[!ht]
	\centering
	\caption{Summary of Major Notations}
	\label{Table1}
	\begin{tabular}{|l|p{10.0cm}|}
		\hline
		{\bf Notation} & { \bf Description} 
		\\ 
		\hline
		$\lambda$ & Base station (BS) intensity (average number of BSs in a unit area)
		\\ 
		\hline
		$N$ & Number of devices in each cell 
		\\ 
		\hline
		$M$ & Number of available resource blocks at each BS
		\\
		\hline
		$F$, $F_k$ & Global loss function, local loss function at device $k$
		\\
		\hline 
		$w$, $w^k$ & Global model parameters, local model parameters at device $k$
		\\ 
		\hline
		$p_k$ & Weight of $k$-th nearest device
		\\
		\hline
		$q_k$ & Average number of allocated resource blocks to device $k$ at a sampling step 
		\\
		\hline
		$U_k$ & Success probability of device $k$  
		\\
		\hline
		$K$ & Total number of FL rounds (iterations)
		\\
		\hline
		$E$ & Number of local SGD steps during each round of FL
		\\
		\hline
		$\ell$ & Number of transmission attempts at each aggregation step
		\\
		\hline
	\end{tabular}
	
\end{table}

\section{System Model, Assumptions, and Proposed FL Algorithm}
\subsection{Network Model}
Consider a single-tier cellular network. The locations of the BSs follow a homogeneous Poisson point process (PPP) $\Phi$ of intensity $\lambda$, and each BS serves the user devices that are located in its Voronoi cell, i.e. each device is associated to its nearest BS. In each cell, $N$ devices are uniformly distributed; we use subscript $k$ to denote the $k$-th nearest device to the BS. Thus, $r_{k}$ denotes the distance between the serving BS and its $k$-th nearest device. Each BS allocates $M$ resource blocks for the learning process, where $M \le N$.
 
\subsection{Federated Learning}
In order to learn a statistical model from the distributed data across user devices, BS tries to solve the following distributed optimization problem:
\begin{IEEEeqnarray}{rCl}
	\min_w  \qquad F(w)=\sum_{k=1}^{N} p_k F_k(w),
	\label{eq:Global_loss}
\end{IEEEeqnarray}
where $w$ is the learning model parameters. $p_k$ is the weight of the $k$-th nearest device such that $p_k \ge 0$ and $\sum_{k=1}^{N} p_k=1$. $F_k(w)$ also denotes the local loss function at device $k$; it is defined as
\begin{IEEEeqnarray}{rCl}
	F_k(w) = \frac{1}{n_k} \sum_{x\in \mathcal{D}_k} \mathcal{L}(w,x),
	\label{eq:Local_loss}
\end{IEEEeqnarray}
where $\mathcal{D}_k$ is the local dataset at device $k$, and is non-i.i.d. across different devices. $n_k=|\mathcal{D}_k|$ denotes the number of samples in $\mathcal{D}_k$. Thus, in \eqref{eq:Global_loss}, we can set $p_k=\frac{n_k}{n}$, where $n=\sum_{k=1}^{N}n_k$. $ \mathcal{L}(w,x)$ also represents the loss function for data sample $x$. 

\begin{figure}
	\centering
	\includegraphics[width=.9\textwidth]{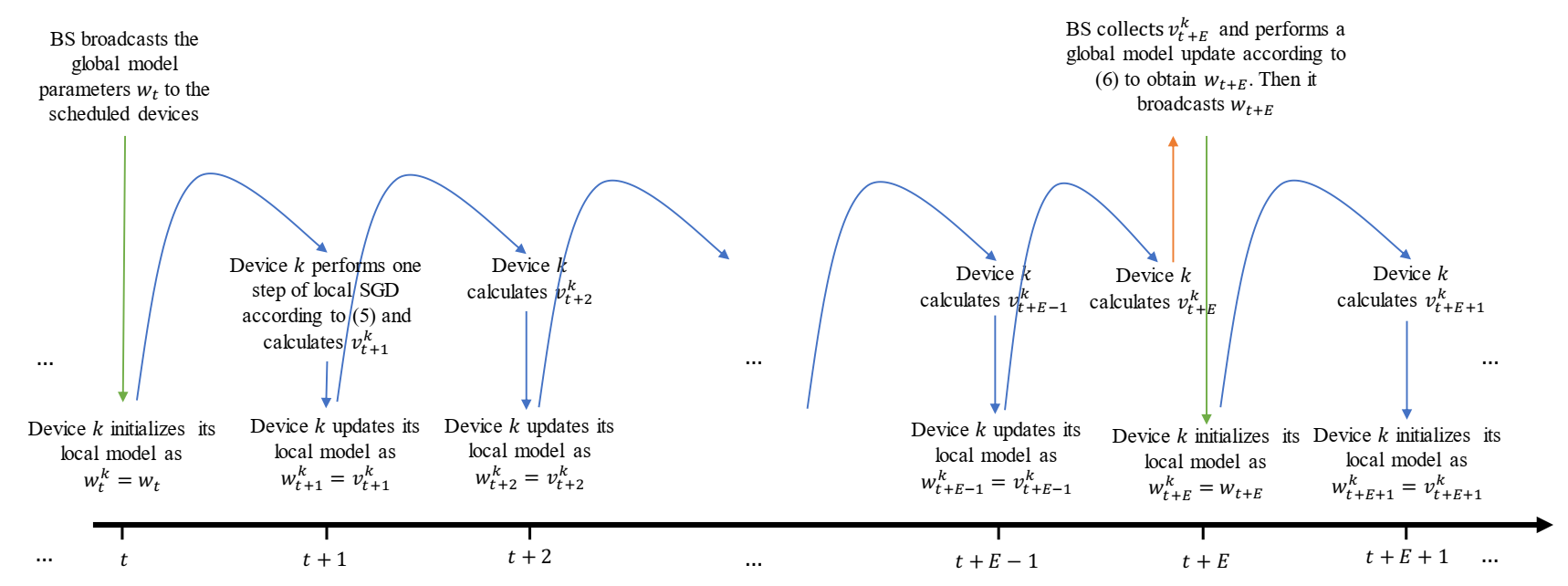}
	\caption{One iteration (round) of the distributed learning algorithm starting at time $t$. Green arrows illustrate the broadcast step. Blue arrows correspond to the local SGD step, and red arrow represents the aggregation step.}
	\label{fig:Algorithm}	
\end{figure}

Since the information is distributed across multiple devices, a BS cannot directly solve \eqref{eq:Global_loss}. Therefore, the BS and the devices collaboratively learn the optimal model parameters $w^*$ \footnote{$w^*=\arg \min F(w)$} by following an iterative algorithm. Specifically, after initializing the model parameters at the BS at time $0$, each iteration (round) of the algorithm comprises: 1) \textit{Sampling and broadcast}, 2) \textit{Local stochastic gradient descent (SGD)}, and 3) \textit{Aggregation and averaging}. In the following, we discuss each of these steps in detail for an iteration that starts at time $t$. This iteration is also shown in \figref{fig:Algorithm}. 

\textbf{Sampling and Broadcasting}: Due to the limited number of available resource blocks, the BS selects a group of devices at time $t$ and broadcasts its model parameters to these devices. In this paper, we consider two different sampling schemes.

In {\em Scheme I}, the BS uniformly selects $M$ devices out of $N$ devices without replacement, i.e. each device uses at most one resource block. Let us denote the set of the selected (scheduled) devices at time $t$ by $\mathcal{S}_t$. For {\em Scheme I}, device $k$ uses $q_k$ resource blocks on average at sampling time $t$, where
\begin{IEEEeqnarray}{rCl}
	q_k = \mathbb{E}\left[ \sum_{m=1}^{M} \mathbf{1}\left(k\in\mathcal{S}_t(m)\right) \right] = \frac{M}{N}.
	\label{eq:q_SchemeI}
\end{IEEEeqnarray} 
$\mathcal{S}_t(m)$ in the above equation denotes the index of the scheduled device at time $t$ for resource block $m$, and $\mathbf{1}(.)$ is the indicator function.

In {\em Scheme II}, at time $t$, for resource block $m$, the BS samples a device with replacement from $N$ devices with probabilities $\{\hat{q}_k\}$. Thus, sampling is independently and identically distributed over $m$, and some devices may use more that one resource block. For Scheme II, we have
\begin{IEEEeqnarray}{rCl}
	q_k = \mathbb{E}\left[ \sum_{m=1}^{M} \mathbf{1}\left(k\in\mathcal{S}_t(m)\right) \right] = M\hat{q}_k.
	\label{eq:q_SchemeII}
\end{IEEEeqnarray} 

\textbf{Local SGD}: After receiving the global model parameters from the BS at time $t$, the scheduled device $k$ initializes its local model as $w_t^k=w_t$, where $w_t$ is the global model parameters sent by the BS, and $w_t^k$ is the local model parameters at device $k$. Then device $k$ performs $E$ steps of SGD and updates its local model at each step as follows:
\begin{IEEEeqnarray}{rCl}
	v_{t+i+1}^k = w_{t+i}^k - \eta_{t+i} \nabla F_k(w_{t+i}^k;\xi_{t+i}^k), \qquad i=0,...,E-1,
	\label{eq:SGD-example}
\end{IEEEeqnarray}
where $\eta_{t+i}$ is the learning rate, and $\xi_{t+i}^k$ is the sample uniformly chosen from the local dataset at device $k$\footnote{In later sections, we will discuss about generalizing our results to batch sizes which are larger than one.}. Moreover,
\begin{IEEEeqnarray}{rCl}
	w_{t+i}^k = 
	\begin{cases}
		w_t       & i = 0 \\
		v_{t+i}^k & i = 1,2,...,E-1,
	\end{cases}. \nonumber
\end{IEEEeqnarray}

\textbf{Aggregation and Averaging}: At the end of $E$ steps of stochastic gradient descent, the scheduled devices transmit $v_{t+E}^k-w_t$ to the BS using the allocated resource block(s). After successfully collecting the  received local models, the BS updates the global model as 
\begin{IEEEeqnarray}{rCl}
	w_{t+E} = w_t + \sum_{k=1}^{N} \sum_{m=1}^{M} \frac{p_k}{q_kU_k} 
	\mathbf{1}\left(k\in \mathcal{S}_t(m), \text{SINR}_{k,m}>\theta\right) 
	(v_{t+E}^k-w_t),
	\label{eq:Averaging-example}
\end{IEEEeqnarray}  
where $\text{SINR}_{k,m}$ denotes the received signal-to-interference-plus-noise ratio (SINR) of device $k$ over resource block $m$. $\theta$ denotes the SINR threshold, and $U_k$ is defined as 
\begin{IEEEeqnarray}{rCl}
	U_k = \mathbb{E}\left[ \mathbf{1}
	\left( \text{SINR}_{k,m}>\theta \right) \mid
	k\in \mathcal{S}_t(m) \right], \nonumber
\end{IEEEeqnarray} 
i.e. $U_k$ denotes the success probability of device $k$ given that it is scheduled to use resource block $m$.

Based on the above discussion, we have summarized our proposed FL algorithm in \textbf{Algorithm~\ref{algo}}. 
\begin{algorithm}
	\DontPrintSemicolon 
	\KwIn{$K$, $E$}
	\kwInit{Global model parameters at the BS $w_0$}
	\For{ $i=0,1,...,K-1$ }{ 
		\textbf{Step 1 (Sampling and Broadcast):} The BS samples a subset of devices and broadcasts the global model parameters $w_{iE}$ to the sampled devices
		
		\textbf{Step 2 (Local SGD):} Each sampled device initializes its local model parameters with $w_{iE}$, and performs $E$ steps of local SGD following \eqref{eq:SGD-example} with $t=iE$
		
		\textbf{Step 3 (Aggregation and Averaging):} Scheduled devices transmit their updated local models to the BS. Then, the BS updates the global model parameters according to \eqref{eq:Averaging-example} and obtains $w_{(i+1)E}$
	}
	\Return{$w_{KE}$}\;
	\caption{Proposed FL Algorithm}
	\label{algo}
\end{algorithm}

Moreover, we can summarize the update rule of the proposed algorithm as
\begin{IEEEeqnarray}{rCl}
	v_{t+1}^k &=& w_t^k - \eta_t \nabla F_k(w_t^k; \xi_t^k), \qquad \forall k\in\{1,...,N\}, 
	\label{eq:SGD}
\end{IEEEeqnarray}
where
\begin{IEEEeqnarray}{rCl}
	w_t^k &=& 
	\begin{cases}
		w_0 & t=0 					  \\
		v_t^k & t\not\in\mathcal{I}_E \\
		w_t = 
		w_{t-E} + \sum_{k=1}^{N} \sum_{m=1}^{M} \frac{p_k}{q_kU_k} \mathbf{1}\left(k\in \mathcal{S}_{t-E }(m), \text{SINR}_{k,m}>\theta\right) (v_t^k-w_{t-E}),
		& t\in\mathcal{I}_E 
	\end{cases}, \nonumber \\
	\label{eq:Averaging}
\end{IEEEeqnarray}
in which $\mathcal{I}_E=\left\{nE \mid n=1,2,...,\right\}$ is the set of time indexes of the global synchronization steps after $t=0$ \cite{li2019convergence}.

\subsection{Interference, SINR, and Success Probability}
In this subsection, we study the success probability of device $k$ at its associated BS, given that it is scheduled to transmit over resource block $m$. Due to the stationarity of the homogeneous PPP \cite{haenggi2012stochastic}, we can consider the location of this BS as the origin of our coordination system.

Before studying the uplink success probability, it is worth mentioning that, in this paper, we  assume that downlink communication is always successful, i.e. all the scheduled devices successfully receive the global model parameters. This is a valid assumption since the BS can transmit with more power compared to the devices; moreover, the BS can allocate more than one resource block for broadcasting the global model parameters. However, due to power constraint at user devices and limited number of allocated resource blocks to each scheduled device, we consider that uplink transmissions may not be successful at an aggregation step.

To increase the success probability during the aggregation steps, we assume that all scheduled devices transmit their local model parameters for $\ell$ times, and the BS employs selection combining, i.e. it uses the highest received SINR to recover the local model parameters transmitted over resource block $m$. Therefore, 
\begin{IEEEeqnarray}{rCl}
	U_k &=& \mathbb{E}\left[ \mathbf{1} \left( \text{SINR}_{k,m}>\theta \right) \mid k\in \mathcal{S}_t(m) \right] \nonumber \\
	    &=& \mathbb{E}\left[ \mathbf{1} \left( \max\left\{\text{SINR}_{k,m}(1),\text{SINR}_{k,m}(2),...,\text{SINR}_{k,m}(\ell) \right\}>\theta \right) \mid k\in \mathcal{S}_t(m) \right], \nonumber 
\end{IEEEeqnarray}
where $\text{SINR}_{k,m}(i)$ denotes the received SINR from device $k$ over resource block $m$ at $i$-th transmission attempt of an aggregation step. 

Since the success event is identically distributed across different iterations (rounds) of the algorithm for device $k$, $\text{SINR}_{k,m}(i)$ is not indexed by the iteration number. Moreover, at the aggregation step of each iteration, interference and fading are identically distributed across different resource blocks; thus, we can omit subscript $m$ from $\text{SINR}_{k,m}(i)$ since its statistics does not depend on $m$. With this in mind, we can write
\begin{IEEEeqnarray}{rCl}
	\text{SINR}_{k}(i) = \frac{h_{k}(i)r_k^{-\alpha}}{I(i) +\sigma^2} 
	                   = \frac{h_{k}(i)r_k^{-\alpha}}{\sum_{x\in\Phi_{\rm I}} h_x(i) \|x\|^{-\alpha}+\sigma^2}, 
	\label{eq:SINR}
\end{IEEEeqnarray}
where $\sigma^2$ is the normalized noise power (noise power to the device transmit power). $I(i) = \sum_{x\in\Phi_{\rm I}} h_x(i) \|x\|^{-\alpha}$ is the interference at $i$-th transmission attempt with $\Phi_{\rm I}$ denoting the set of interferers. Here $h_x(i)$ denotes the 
small-scale fading gain between a device at location $x$ and the BS at the origin. We consider Rayleigh fading, i.e. $h_x\sim\exp(1)$, and it is i.i.d. across $i$ (different transmission attempts) and $x$ (different locations). $\alpha$ denotes the path-loss exponent. Since the set of interferers remains the same during different transmission attempts of one aggregation step,  the success events over one resource block during an aggregation step, are temporally correlated.

It is also worth mentioning that we can easily extend our framework to more complex scenarios including uplink transmission with power control as in \cite{elsawy2014stochastic}, multiple input multiple output antenna systems as in \cite{di2016stochastic,bai2016analyzing}, millimeter wave as in \cite{singh2015tractable}.

\begin{Lemma}
	For the described network, with $\ell$ transmission attempts at each aggregation step, the success probability for scheduled device $k$ is
	\begin{IEEEeqnarray}{rCl}
		U_k \approx \sum_{i=1}^{\ell} \binom{\ell}{i} (-1)^{i+1} 
		\exp\left\{-i\theta\sigma^2r_k^{\alpha}
		-2\pi\lambda \int_{0}^{\infty} \left(1-\frac{1}{ \left(1+\theta r_k^{\alpha}x^{-\alpha}\right)^i } \right) (1-e^{-12/5\lambda\pi x^2}) x {\rm d}x \right\}.
		\nonumber
	\end{IEEEeqnarray}  
\end{Lemma}
\begin{IEEEproof}
	See \textbf{Appendix A}.
\end{IEEEproof}

\section{Convergence Analysis of the Proposed Federated Learning Algorithm}

\subsection{Convergence Rate}
In this section, we prove the convergence of the proposed algorithm. In this regard, we first introduce additional notations and assumptions that are required to derive the convergence rate. 

Following the same notation as in \cite{li2019convergence}, for time $t$, we define 
\begin{itemize}
	\item $\bar{v}_t=\sum_{k=1}^{N}p_k v_t^k$,
	\item $\bar{w}_t=\sum_{k=1}^{N}p_k w_t^k$,
	\item $g_t = \sum_{k=1}^{N} p_k \nabla F_k(w_t^k;\xi_t^k)$,
	\item $\bar{g}_t = \sum_{k=1}^{N} p_k \nabla F_k(w_t^k)$, where $\nabla F_k(w_t^k) = \mathbb{E}_{\xi}\left[\nabla F_k(w_t^k;\xi) \right]$.
\end{itemize}
Thus, at $t\in\mathcal{I}_E$, $w_t^k=\bar{w}_t=w_t$. 

We also make the following assumptions \cite{li2019convergence,zhang2013communication,stich2018local}.

\begin{Assumption}
	$F_1,...,F_N$ are all $\mu$-strongly convex, i.e. for $k\in\{1,...,N\}$, 
	\begin{IEEEeqnarray}{rCl}
		F_k(w_2) \ge F_k(w_1) + \nabla F_k(w_1)^T(w_2-w_1) + \frac{\mu}{2} \|w_2-w_1\|^2, \qquad \forall w_1, w_2. \nonumber
	\end{IEEEeqnarray}
	Consequently, $F$ is $\mu$-strongly convex.
\end{Assumption}
\begin{Assumption}
	$F_1,...,F_N$ are all $L$-smooth, i.e. for $k\in\{1,...,N\}$,
		\begin{IEEEeqnarray}{rCl}
		F_k(w_2) \le F_k(w_1) + \nabla F_k(w_1)^T(w_2-w_1) + \frac{L}{2} \|w_2-w_1\|^2, \qquad \forall w_1, w_2. \nonumber
	\end{IEEEeqnarray}
	Consequently, $F$ is $L$-smooth.
\end{Assumption}
\begin{Assumption}
	The variance of stochastic gradient at device $k$, $k\in\{1,...,N\}$, is upper bounded by $\sigma_k^2$, i.e.,
	\begin{IEEEeqnarray}{rCl}
		\mathbb{E}_{\xi}\left[ \| \nabla F_k(w;\xi) - \nabla F_k(w) \|^2\right] \le \sigma_k^2, \qquad \forall w. \nonumber
	\end{IEEEeqnarray}
\end{Assumption}
\begin{Assumption}
	The second moment of the norm of the stochastic gradient is bounded at all devices. For all $k\in\{1,...,N\}$, 
	\begin{IEEEeqnarray}{rCl}
		\mathbb{E}_{\xi} \left[ \| \nabla F_k(w;\xi) \|^2 \right] \le G^2, \qquad \forall w. \nonumber
	\end{IEEEeqnarray}
\end{Assumption}
We also define $\Gamma=F^*-\sum_{k=1}^{N}F_k^*$, where $F^*$ is the optimal value (minimum) of the objective function in \eqref{eq:Global_loss} and $F_k^*$ is the optimal value (minimum) of the objective function in \eqref{eq:Local_loss}. $\Gamma \ge 0$, and increases as the heterogeneity (degree of non-i.i.d.) of the data distribution increases \cite{li2019convergence}.

\begin{Theorem}
	When $\eta_t=\frac{2}{\mu(\gamma+t)}$ with $\gamma=\max\left\{8\frac{L}{\mu},E\right\}$, after the averaging step at time $T$, we have
	\begin{multline}
		\mathbb{E}\left[ F(w_T)-F^* \right] \le 
		\\
		\frac{L/\mu}{\gamma+T} 	\left[ 
		\frac{2}{\mu}\left( \sum_{k=1}^{N} p_k^2\sigma_k^2 + 6L\Gamma + 8(E-1)^2G^2 + 4E^2G^2B \right)+ 
		\frac{\mu \gamma}{2} \left\| w_{0}-w^* \right\|^2 \right], \nonumber
	\end{multline}
	where $B=\sum_{k=1}^N p_k \left( \frac{1}{q_k U_k}-1 \right)$ for sampling Scheme I, and $B=\sum_{k=1}^N  p_k\left( \frac{1}{q_k U_k}-\frac{1}{M} \right)$ for sampling Scheme II. $w_0$ denotes the initialized global model parameters.
\end{Theorem}
\begin{IEEEproof}
	See \textbf{Appendix B}.
\end{IEEEproof}
Therefore, when \textbf{Assumptions 1 to 4} hold, the proposed federated learning algorithm converges with rate $\mathcal{O}\left( \frac{1}{T} \right)$. So far we have assumed that each device uses only one data sample at each local update step. In the following, we discuss using mini-batches with more than one data sample at the local update steps.

\subsection{Mini-Batch Gradient Descent}
When we use mini-batches with size $b$ for local update steps, \eqref{eq:SGD} changes to 
\begin{IEEEeqnarray}{rCl}
	v_{t+1}^k &=& w_t^k - \eta_t \nabla F_k(w_t^k; \{\xi_t^k\}),
	\qquad \forall k\in\{1,...,N\}, 
	\label{eq:mini-batch-SGD}
\end{IEEEeqnarray} 
where 
\begin{IEEEeqnarray}{rCl}
	\nabla F_k(w_t^k; \{\xi_t^k\}) = \frac{1}{b}\sum_{\xi\in\mathcal{B}_t^k} \nabla F_k(w_t^k; \xi) \nonumber
\end{IEEEeqnarray} 	
is the estimated gradient at device $k$ at time $t$ using samples in the mini-batch $\mathcal{B}_t^k=\{\xi_t^k\}$. For device $k$ and $\forall w$, we have 
\begin{IEEEeqnarray}{rCl}
	\mathbb{E}_{\left\{\xi\right\}} \left[ \left\| \nabla F_k(w; \{\xi\}) - \nabla F_k(w) \right\|^2\right] &=& 
	\mathbb{E}_{\left\{\xi\right\}} \left[ \left\| \sum_{\xi\in\mathcal{B}^k} \frac{1}{b}\left( \nabla F_k(w; \xi) - \nabla F_k(w) \right) \right\|^2\right] 
	\nonumber \\
	&\le&
	\mathbb{E}_{\left\{\xi\right\}} \left[ \sum_{\xi\in\mathcal{B}^k} \frac{1}{b} \left\| \left( \nabla F_k(w; \xi) - \nabla F_k(w) \right) \right\|^2\right]
	\stackrel{\text{(a)}}{\le} \sigma_k^2, \nonumber
\end{IEEEeqnarray}
where (a) is obtained from \textbf{Assumption 3}. Similarly, from \textbf{Assumption 4}, we have
\begin{IEEEeqnarray}{rCl}
	\mathbb{E}_{\left\{\xi\right\}} \left[ \left\| \nabla F_k(w; \{\xi\}) \right\|^2 \right] \le G^2, \qquad \forall w. \nonumber
\end{IEEEeqnarray}
Therefore, \textbf{Theorem 1} holds for any batch size.

\section{Further Analysis and Comparison}
In this section, we provide a further discussion on the proposed federated learning algorithm using \textbf{Theorem 1}. Specifically, we first study the impact of number of iterations (rounds), number of local update steps at each round, and number of transmission attempts at each aggregation step in each round; then we find the optimal scheduling policy that maximizes the convergence rate. Finally, we compare our algorithm with the works that do not consider the transmission success probability and prove its significance. 

\subsection{Effects of Number of Computation and Communication Steps}
To study the impact of number of rounds ($K$\footnote{According to the definition, $K=T/E$ in \textbf{Theorem 1}.}), number of local update steps during each round ($E$), and number of transmission attempts at each aggregation step in each round ($\ell$), we use a simpler form of \textbf{Theorem 1}. 
\begin{Corollary}
	When $\eta_t=\frac{2}{\mu(\gamma+t)}$ with $\gamma=\max\left\{8\frac{L}{\mu},E\right\}$, after the averaging step at the $K$-th round, we have
	\begin{multline}
	\mathbb{E}\left[ F(w_T)-F^* \right] \le 
	\\
	\frac{L/\mu}{KE} 	\left[ 
	\frac{2}{\mu}\left( \sum_{k=1}^{N} p_k^2\sigma_k^2 + 6L\Gamma + 8(E-1)^2G^2 + 4E^2G^2B \right)+ 
	\frac{\mu \gamma}{2} \left\| w_{0}-w^* \right\|^2 \right], \nonumber
	\end{multline}
	where $T=KE$ and $B$ is defined in \textbf{Theorem 1}.
\end{Corollary}

By taking the derivative of the right hand side with respect to $E$, we observe that the upper bound in \textbf{Corollary 1} first decreases and then increases; thus, an optimal value for $E$ exists. In fact this can be easily understood by considering that increasing $E$ at first allows each device to move further in the direction of the optimal model parameters; thus, increases the convergence. However, when $E$ is set to a large value, each device moves towards its local optimum model instead of the global optimum.

It is also obvious that the upper bound in \textbf{Corollary 1} is a decreasing function of $K$ and $\ell$, i.e. the algorithm converges faster as we increase the number of communications. However, increasing $K$ increases the rate of convergence more than increasing $\ell$. To prove this statement, we denote the success probability after $\ell$ transmission attempts by $U_k^{(\ell)}$. According to \eqref{eq:A-mid_step},
\begin{IEEEeqnarray}{rCl}
	U_k^{(\ell)} &=& 1-\mathbb{E}_{\Phi_{\rm I}}\left[ 
	\left( 1-e^{-\theta\sigma^2r_k^{\alpha}} \prod_{x\in\Phi_{\rm I}} \frac{1}{1+\theta r_k^{\alpha}\|x\|^{-\alpha}} \right)^\ell 
	\right] \nonumber \\
	&\stackrel{\text{(a)}}{\le}& 1-\mathbb{E}_{\Phi_{\rm I}}\left[ 
	1-\ell  e^{-\theta\sigma^2r_k^{\alpha}} \prod_{x\in\Phi_{\rm I}} \frac{1}{1+\theta r_k^{\alpha}\|x\|^{-\alpha}} 
	\right] \nonumber \\
	&\le& \ell U_k^{(1)}, \label{eq:U_k^l}
\end{IEEEeqnarray} 
where, in (a), we have used Bernoulli's inequality \cite{mitrinovic1970analytic}. When we increase $K$ to $\ell K$, from the upper bound in \textbf{Corollary 1} we observe
\begin{IEEEeqnarray}{rCl}
	\frac{L/\mu}{\ell KE} 	\left[  \textbf{Constant}_1 + \textbf{Constant}_2 \times \sum_{k=1}^{N}\frac{p_k}{q_kU_k^{(1)}} \right]
	&\le& 
	\frac{L/\mu}{KE} 	\left[  \textbf{Constant}_1 + \textbf{Constant}_2 \times \sum_{k=1}^{N}\frac{p_k}{\ell q_kU_k^{(1)}} \right]
	\nonumber \\
	&\stackrel{\text(a)}{\le}& 
	\frac{L/\mu}{KE} 	\left[  \textbf{Constant}_1 + \textbf{Constant}_2 \times \sum_{k=1}^{N}\frac{p_k}{q_kU_k^{(\ell)}} \right],
	\nonumber
\end{IEEEeqnarray}
where we have used \eqref{eq:U_k^l} in (a). Although increasing $K$ is more powerful compared to increasing $\ell$ in improving the convergence rate, it consumes more resources since, during each round of the learning process, the scheduled devices perform $E$ steps of local update.

\subsection{Scheduling Policy}
In this subsection, we find the optimal scheduling policy for Scheme II. In this regard, we first find the solution to the following optimization problem:
\begin{mini}|r|
	{\{q_k\}}  {\mathbb{E}\left[F(w_T)\right]}{}{} 
	\addConstraint {\sum_{k=1}^{N}q_k}{=M}{} 
	\addConstraint {q_k}{>0, \qquad}{\forall k\in\{1,...,N\}.}
	\label{eq:P0}
\end{mini}  
The first condition is due to the fact that the BS allocates $M$ resource blocks for the learning process. Using the upper bound in \textbf{Theorem 1} as an approximation of $\mathbb{E}\left[F(w_T)-F^*\right]$, we can obtain the solution to \eqref{eq:P0} by solving
\begin{mini}|r|
	{\{q_k\}}  {\sum_{k=1}^{N}\frac{p_k}{U_k}q_k^{-1}}{}{} 
	\addConstraint {\sum_{k=1}^{N}q_k}{=M}{} 
	\addConstraint {q_k}{>0, \qquad}{\forall k\in\{1,...,N\}.}
	\label{eq:P1}
\end{mini}
Now, consider the following optimization
\begin{mini}|r|
	{\{q_k\}}  {\sum_{k=1}^{N}\frac{p_k}{U_k}q_k^{-1}}{}{} 
	\addConstraint {\sum_{k=1}^{N}q_k}{\le M}   {} 
	\addConstraint {q_k}{>0, \qquad}{\forall k\in\{1,...,N\}.}
	\label{eq:P2}
\end{mini}  
By contradiction, one can easily prove that solution to \eqref{eq:P2}, denoted by $q_1^*,...,q_N^*$, satisfies $\sum_{k=1}^{N}q^*_k=M$. Thus, \eqref{eq:P2} is equivalent to \eqref{eq:P1}, i.e. $q_1^*,...,q_N^*$ is also the solution to \eqref{eq:P1}. \eqref{eq:P2} is a geometric program, and it can be transformed to a convex problem \cite{boyd2004convex}. Therefore, we can find $q_1^*,...,q_N^*$ using CVX or any other convex solver~\cite{grant2014cvx}. Finally, the optimal policy for {\em Scheme II} is achieved when $\hat{q}_k=\frac{q_k^*}{M}$. Note that the optimal policy for {\em Scheme II} does not necessarily perform better than {\em Scheme I} since convergence rate and the upper bound in \textbf{Theorem 1} for {\em Scheme I} are different from {\em Scheme II}. In fact, it is only guaranteed that, with optimal policy for Scheme II, the upper bound in \textbf{Theorem 1} performs better compared to other {\em Scheme II} scheduling policies.

\textit{Remark:} {\em Scheme I} samples uniformly without replacement which yields $q_k=\frac{M}{N}$ for this scheme. Nevertheless, \textbf{Theorem 1} is given for any value of $q_k$ for Scheme I. Thus, we can derive $q_1^*,...,q_N^*$ for this scheme. Actually, $q_1^*,...,q_N^*$ for {\em Scheme I} and {\em Scheme II} are identical. However, unlike {\em Scheme II}, it is not possible to calculate the probabilities for {\em Scheme I} from $q_1^*,...,q_N^*$. 

Finally, it is worth mentioning that, according to \textbf{Lemma 3}, the optimal scheduling policy minimizes the variance of the updated global model at each averaging step.

\subsection{Comparison with the Existing Works}
Existing works, such as \cite{mcmahan2017communication, li2020federated, li2019convergence, karimireddy2020scaffold}, assume that the BS always successfully recovers the transmitted information of the scheduled devices. For example, \cite{li2020federated,li2019convergence,nguyen2020fast} use sampling {\em Scheme II} with probabilities $\{p_k\}$ besides the following averaging approach:
\begin{IEEEeqnarray}{rCl}
	w_{t} = \frac{1}{M} \sum_{k=1}^{N}\sum_{m=1}^M\mathbf{1}\left( k\in\mathcal{S}_{t-E}(m) \right) v_t^k \nonumber
\end{IEEEeqnarray}
at $t\in\mathcal{I}_E$. To study this averaging approach when unsuccessful transmission probability is greater than zero, we modify it as
\begin{IEEEeqnarray}{rCl}
	w_{t} = 
	\begin{cases}
		w_{t-E} 
		\\
		\qquad \qquad \qquad \text{if } \sum_{k=1}^{N}\sum_{m=1}^M\mathbf{1}\left( k\in\mathcal{S}_{t-E}(m), \text{SINR}_{k,m}>\theta \right)=0 
		\\
		\frac{1}{M} \sum_{k=1}^{N}\sum_{m=1}^M\mathbf{1}\left( k\in\mathcal{S}_{t-E}(m), \text{SINR}_{k,m}>\theta \right) v_t^k 
		\\
		\qquad \qquad \qquad \text{otherwise }
	\end{cases}. \label{eq:Compare-Averaging1}	
\end{IEEEeqnarray}
In the following, we prove that this averaging scheme does not solve \eqref{eq:Global_loss} when unsuccessful transmission probability is greater than zero and varies across the devices. In this regard, we focus on the following averaging approach:
\begin{IEEEeqnarray}{rCl}
	w_{t} = w_{t-E} + \sum_{k=1}^{N}\sum_{m=1}^M 
	\frac{ \mathbf{1}\left( k\in\mathcal{S}_{t-E}(m), \text{SINR}_{k,m}>\theta \right) }
	     {\sum_{k=1}^{N}\sum_{m=1}^M \mathbf{1}\left( k\in\mathcal{S}_{t-E}(m), \text{SINR}_{k,m}>\theta \right) } \left(v_t^k - w_{t-E}  \right), \qquad t\in\mathcal{I}_E,
    \nonumber \\
	\label{eq:Compare-Averaging2}
\end{IEEEeqnarray}
where we define $\frac{0}{0}=0$. It is straightforward to show that, when $M=1$, \eqref{eq:Compare-Averaging2} is same as \eqref{eq:Compare-Averaging1}.

In \textbf{Appendix C}, we prove that by using \eqref{eq:Compare-Averaging2} for updating the global model parameters at $t\in\mathcal{I}_E$, instead of \eqref{eq:Global_loss}, the algorithm converges to the solution to the following problem:
\begin{IEEEeqnarray}{rCl}
	\min_w  \qquad \hat{F}(w)=\sum_{k=1}^{N} p_kU_k F_k(w).
	\label{eq:Global_loss_distorted}
\end{IEEEeqnarray}

Since \eqref{eq:Compare-Averaging2} and \eqref{eq:Compare-Averaging1} are equal for $M=1$, we can conclude that the employed learning approach by \cite{li2020federated,li2019convergence,nguyen2020fast} do not always solve \eqref{eq:Global_loss}. In fact, one can easily understand that \eqref{eq:Compare-Averaging1} is biased towards devices with higher success probabilities.

\section{Simulation Results and Discussion}

\textbf{Cellular Network}:
BS intensity $\lambda$ is 0.001 (points/area). We assume in each cell there are $N=100$ user devices and each BS allocates only $M=20$ resource blocks for the distributed learning process. We also set $\sigma^2=10^{-4}$, $\alpha=4$, and $\theta=-15\,{\rm dB}$.

\textbf{Datasets}:
We evaluate our proposed federated learning on both real and synthetic datasets. For real data, we use MNIST \cite{lecun1998gradient} and distribute our data similar to \cite{li2019convergence} in a non-i.i.d. fashion. Specifically, each sample in MNIST dataset is a 28$\times$28 image of a handwritten digit between 0 to 9. We distribute the dataset samples such that each device has samples of only two digits and the number of samples at different devices is different. For synthetic data, we follow the same setup as in \cite{shamir2014communication,li2019convergence,li2020federated}. In this regard, we generate samples at device $k$, denoted by $(X_k,Y_k)$, using $y=\text{argmax}\left(\text{softmax}\left( W_k x+b_k \right)\right)$, where $W_k\in\mathbb{R}^{10\times60}$ and $b_k\in\mathbb{R}^{10}$. Each element in $W_k$ and $b_k$ is a realization of $\mathcal{N}(\mu_k,1)$, where $\mu_k \sim \mathcal{N}(0,\tilde{\alpha})$. Thus, $\tilde{\alpha}$ controls the degree of difference between local models at different devices\footnote{When $\tilde{\alpha}=0$, elements of $W_k$ and $b_k$ at all devices are drawn from the same distribution, i.e. $\mathcal{N}(0,1)$.} \cite{li2020federated}. Moreover, $x\in\mathbb{R}^{60}$. The $j$-th element (feature) in $x$ is drawn from $\mathcal{N}(v_{k,j},j^{-1.2})$, where $v_{k,j}\sim\mathcal{N}(B_k,1)$\footnote{$v_{k,j}$ remains the same across $j$-th feature of different samples of device $k$.} and $B_k\sim\mathcal{N}(0,\tilde{\beta})$. Thus, $\tilde{\beta}$ controls the data heterogeneity. Finally, the number of data samples at device $k$, denoted by $n_k$, follows a power law distribution. In this paper, we set $\tilde{\alpha}=1$ and $\tilde{\beta}=1$.

\textbf{Model}: To examine the performance of the proposed FL algorithm on the discussed datasets, we use a three layer neural network with 300 hidden units at each hidden layer. For MNIST datatset, input of the model is a flattened 784-dimensional image. For the synthetic dataset, an input of the model has 60 dimensions. We use ReLU (rectified linear unit) activation function for the hidden layers and softmax in the output layer. At the beginning of round $k$,  $k\in\{0,1,2,...,K-1\}$, we set the learning rate as $\eta_t=\frac{\eta_0}{1+k}$, where $kE \le t < (k+1)E$. We measure the performance of our model with regularized cross-entropy loss, where we use $\ell_2$-norm regularization with regularization parameter $10^{-4}$. 

\begin{figure}
	\parbox[c]{.48\textwidth}{%
		\centerline{\subfigure[Global loss.]
			{\includegraphics[width=0.48\textwidth]{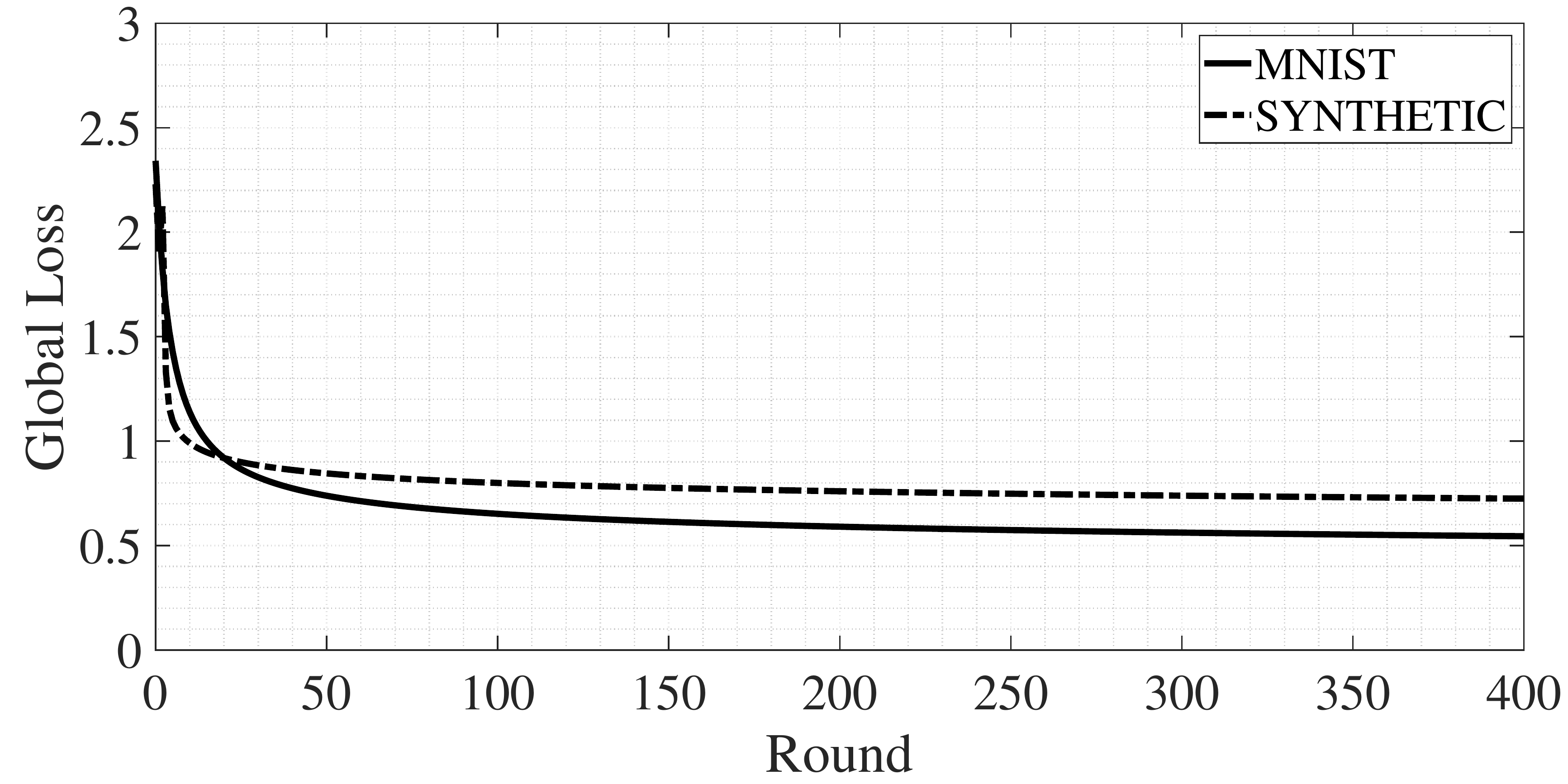}}
			}} 
	\parbox[c]{.48\textwidth}{%
		\centerline{\subfigure[Accuracy.]
			{\includegraphics[width=0.48\textwidth]{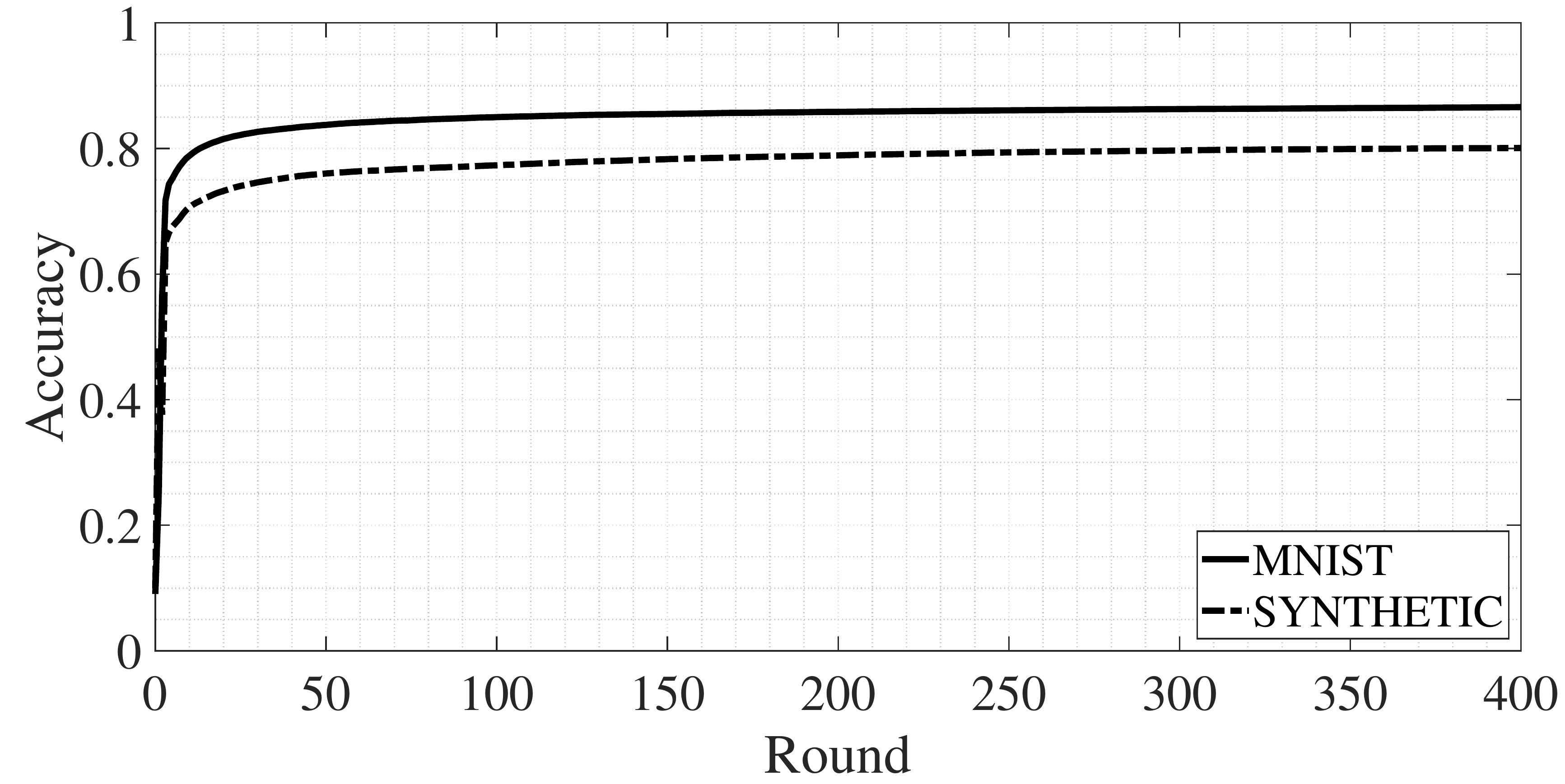}}
			}} 
			
	\caption{Performance of the benchmark algorithm over MNIST and Synthetic datasets.}
	\label{fig:Benchmark}
\end{figure}

\textbf{Benchmark}: When $M=N$ and $U_k=1$, $\forall k\{1,...,N\}$, the BS receives the local model parameters of all devices successfully. Under this assumption, if all devices use their full batch and $E=1$, the distributed learning algorithm is equivalent to the centralized full batch gradient descent. This scenario can be regarded as a benchmark for our proposed algorithm. In \figref{fig:Benchmark}, performance of the centralized full gradient descent with $\eta_0=1$ is shown in terms of global loss and accuracy over both MNIST and synthetic datasets. 

\textbf{Training}: To train the local models at local SGD steps, for MNIST dataset, we use batch sizes of 64 and set $\eta_0=1$. For the synthetic dataset, we use batch sizes of 25 and set $\eta_0=0.1$. Since we use a smaller learning rate for the synthetic dataset, we consider higher values for $E$ at the time of simulation. Our codes are available at \cite{mhdslh2020repo}.

\begin{figure}
	\centering     
	\subfigure[MNIST.]{\includegraphics[width=0.48\textwidth]{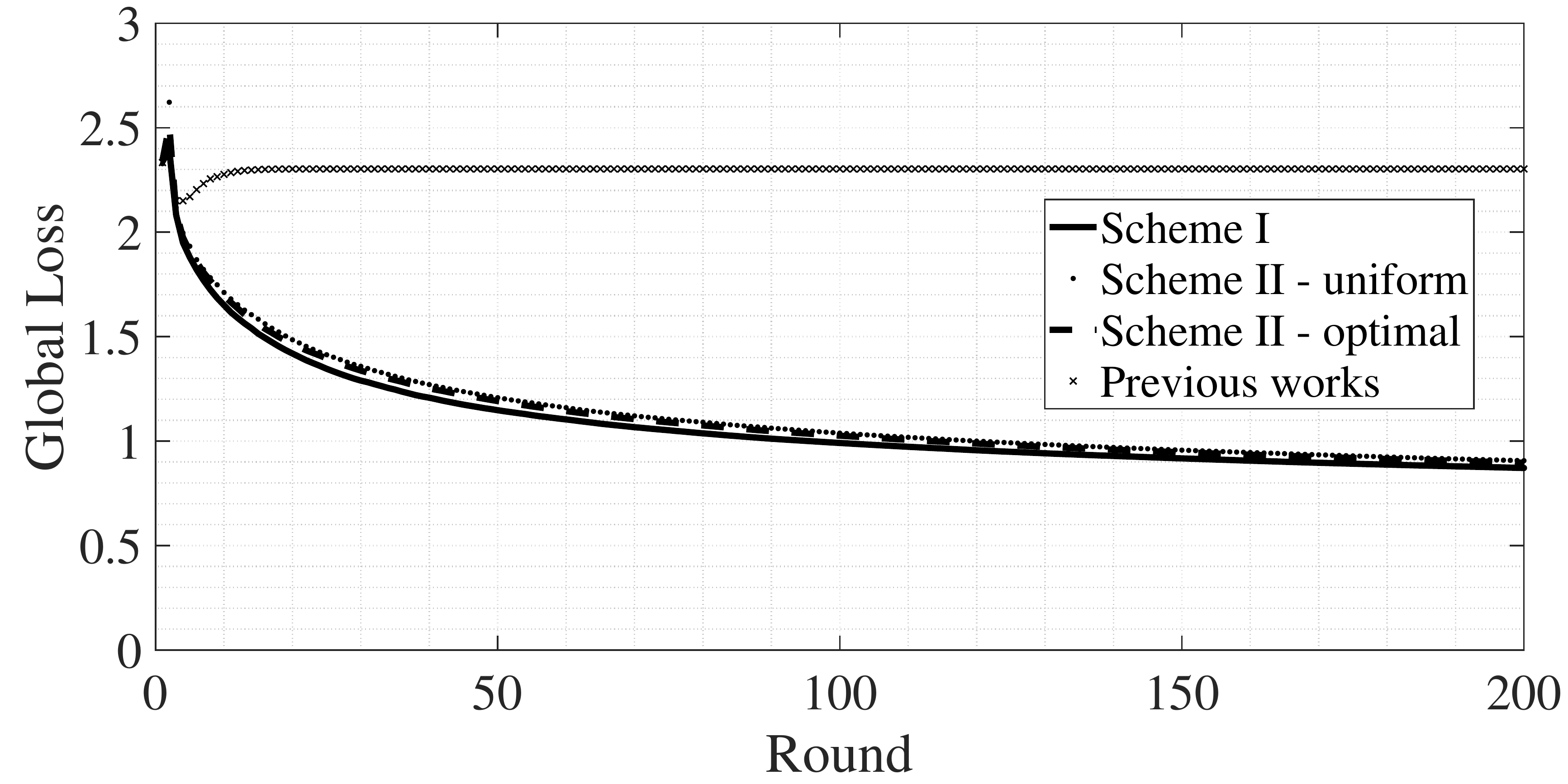}}
	\subfigure[MNIST.]{\includegraphics[width=0.48\textwidth]{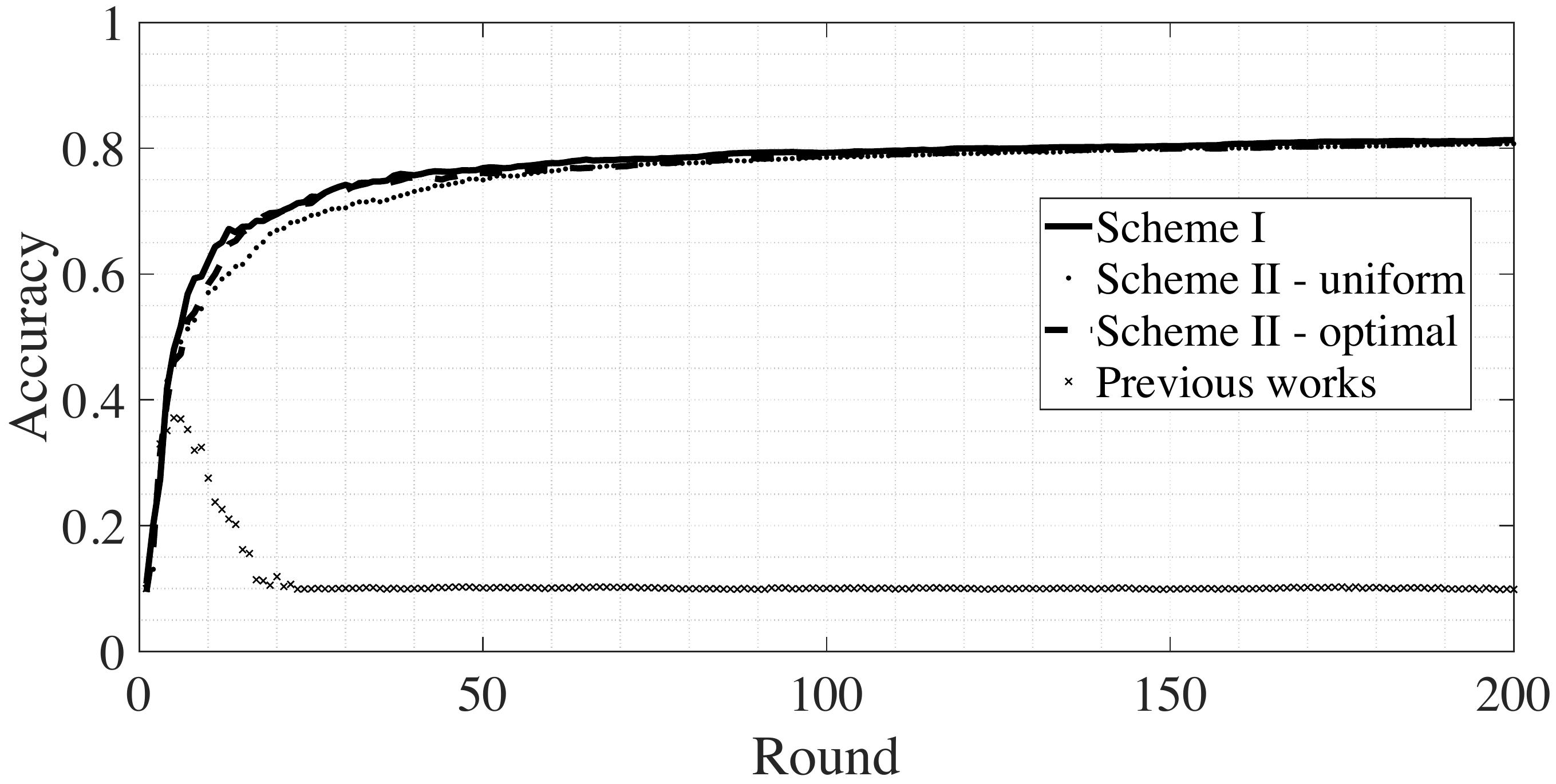}}
	\subfigure[Sythetic.]{\includegraphics[width=0.48\textwidth]{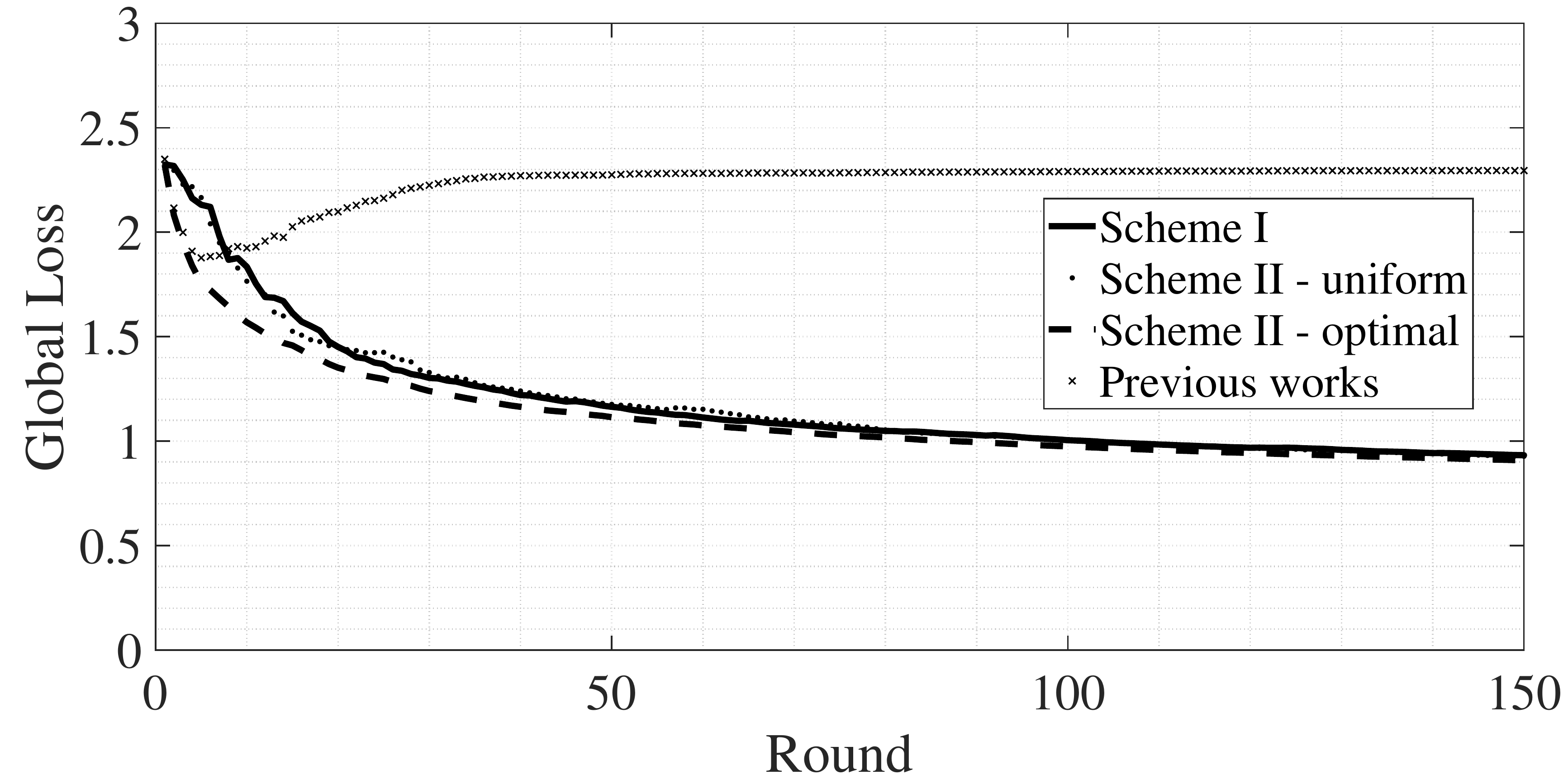}}
	\subfigure[Synthetic.]{\includegraphics[width=0.48\textwidth]{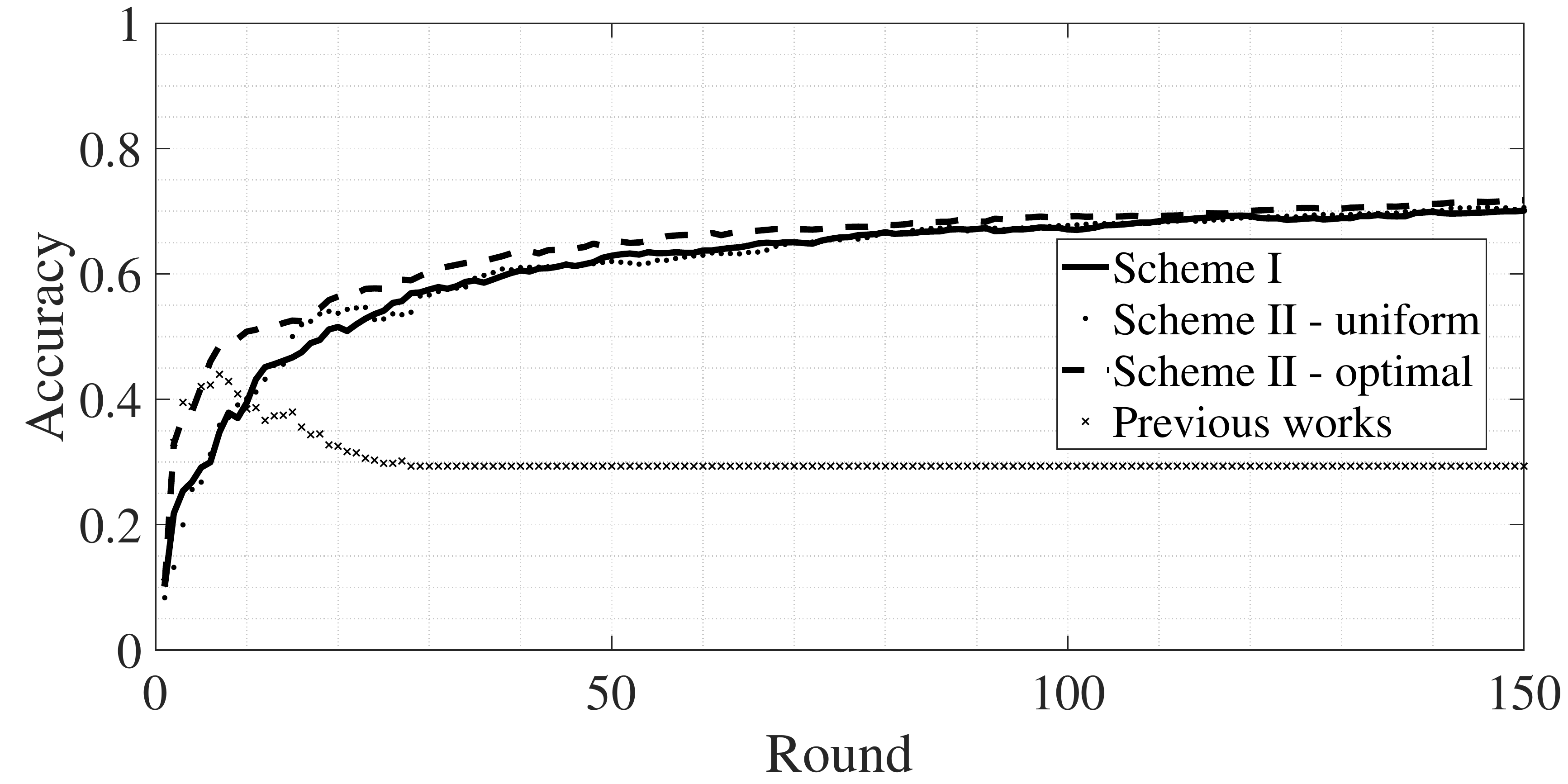}}
	\caption{In (a) and (b), $\ell=2$ and $E=1$. In (c) and (d), $\ell=2$ and $E=20$.}
	\label{fig:Main}
\end{figure}

\textbf{Results\footnote{To make a fair comparison, we have illustrated the average performance over multiple trials.}}: In \figref{fig:Main}(a) and (b), we measure the performance of the proposed FL algorithm in terms of global loss and accuracy on MNIST dataset; \figref{fig:Main}(c) and (d) also illustrate the performance on the synthetic dataset. For {\em Scheme II}, we show the results with $\hat{q}_k=\frac{1}{N}$, i.e. uniform selection and $\hat{q}_k=\frac{q^*_k}{M}$, i.e. optimal selection. For MNIST dataset, the objective function of \eqref{eq:P2} at $q_1=q_2=...=q_N=\frac{M}{N}$ is 7.09 and at the optimal point $q_1^*,q_2^*,...,q_N^*$ is 6.48. For the synthetic dataset, these values are 6.89 and 3.66, respectively. Thus, optimal scheduling has more effect on the synthetic dataset which can also be understood from \figref{fig:Main}. By calculating $B$ in \textbf{Theorem 1} for Scheme I and Scheme II with optimal scheduling, we expect that {\em Scheme I} performs better on MNIST dataset and {\em Scheme II} performs better on the synthetic dataset. This is also illustrated is \figref{fig:Main} and is in compliance with our discussion in \textbf{Section IV.B}. Moreover, as explained in \textbf{Section IV.C}, previous works, such as \cite{li2020federated,li2019convergence,nguyen2020fast}, cannot be used for wireless networks since they do not include unsuccessful transmission probability. This is also shown in \figref{fig:Main}, where we  illustrate the performance of the introduced setup in \textbf{Section IV.C}.

\begin{figure}
	\centering     
	\subfigure[Global loss.]{\includegraphics[width=0.45\textwidth]{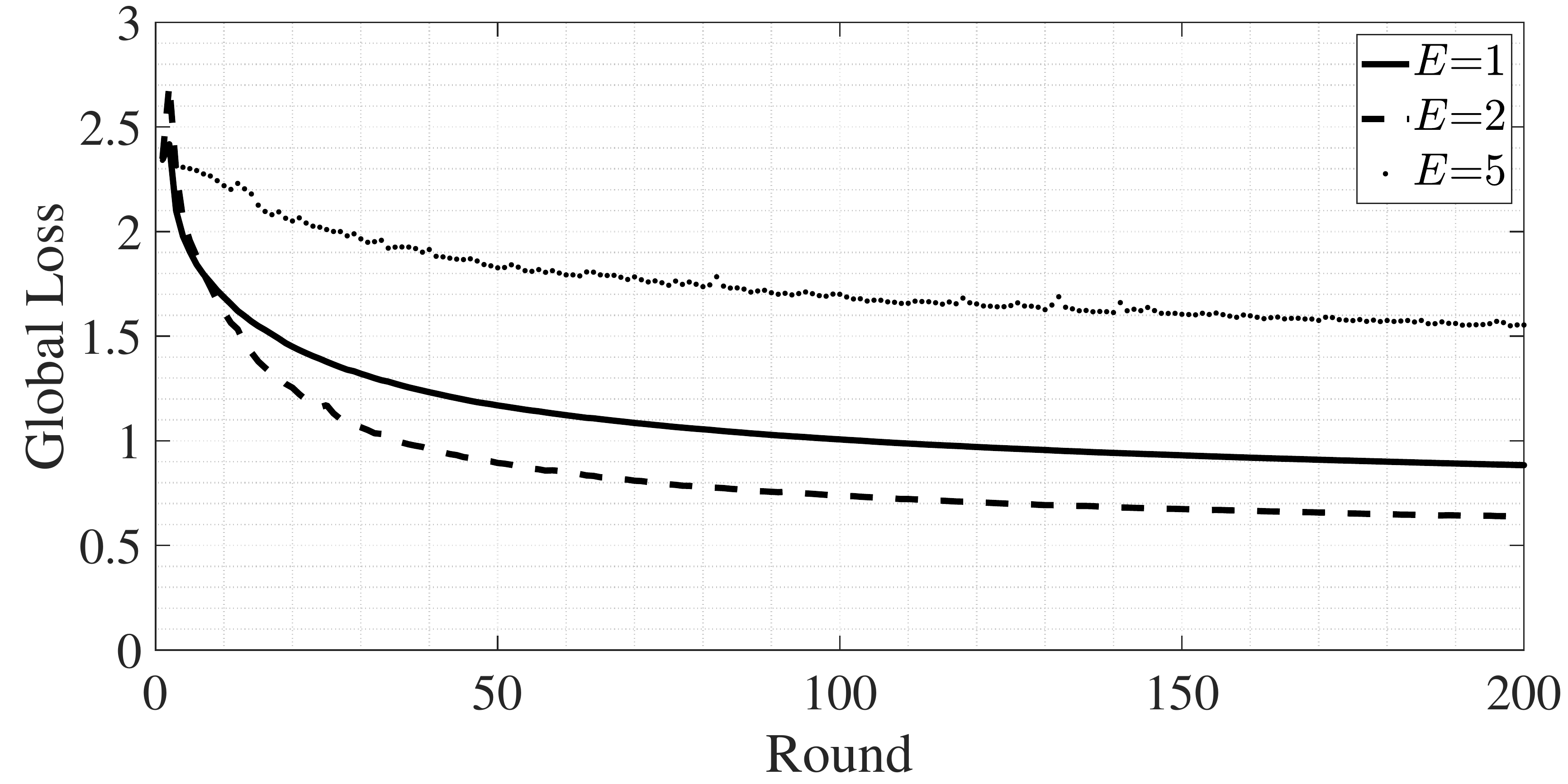}}
	\subfigure[Accuracy.]{\includegraphics[width=0.45\textwidth]{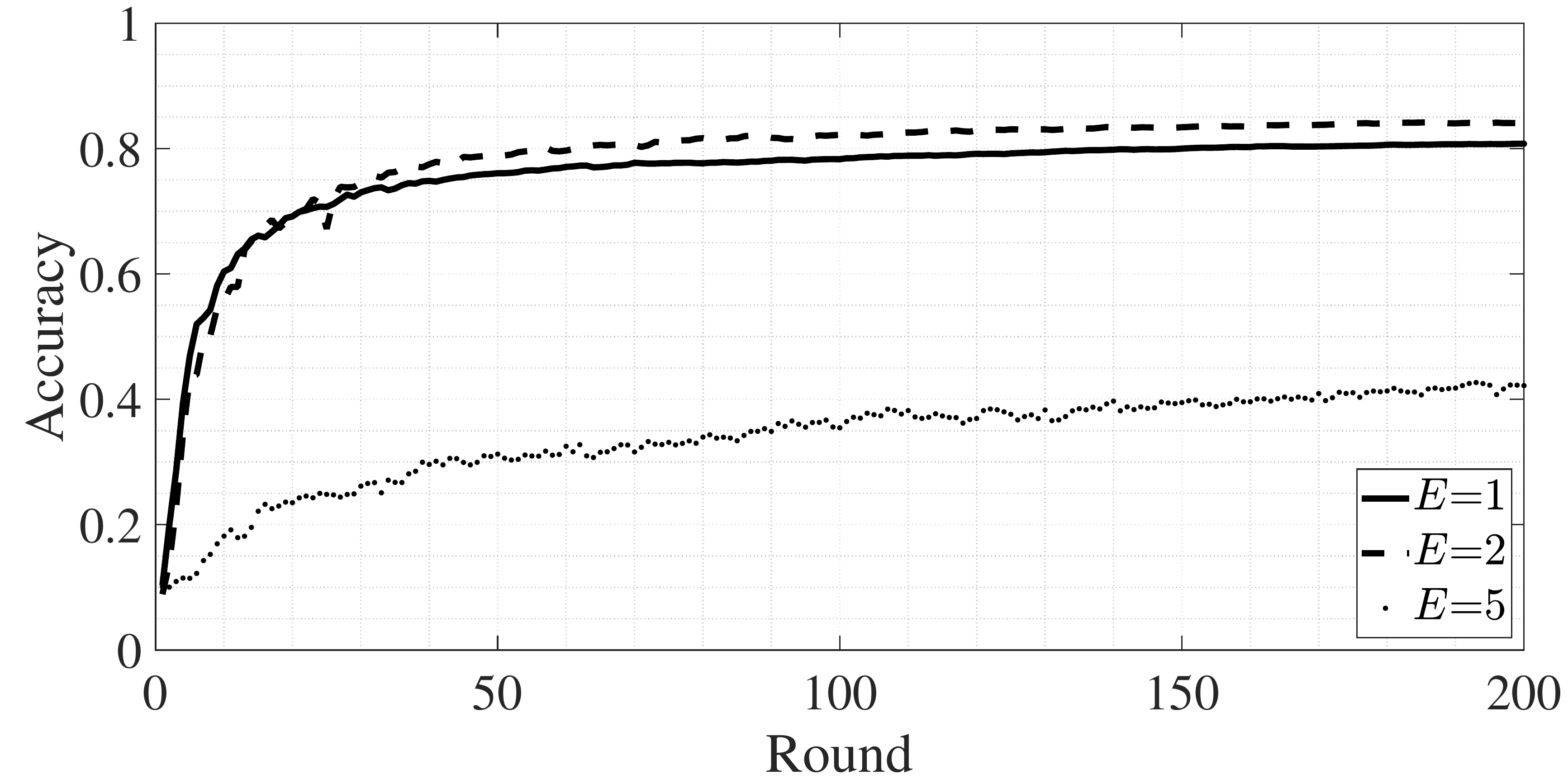}}
	\subfigure[Global loss.]{\includegraphics[width=0.45\textwidth]{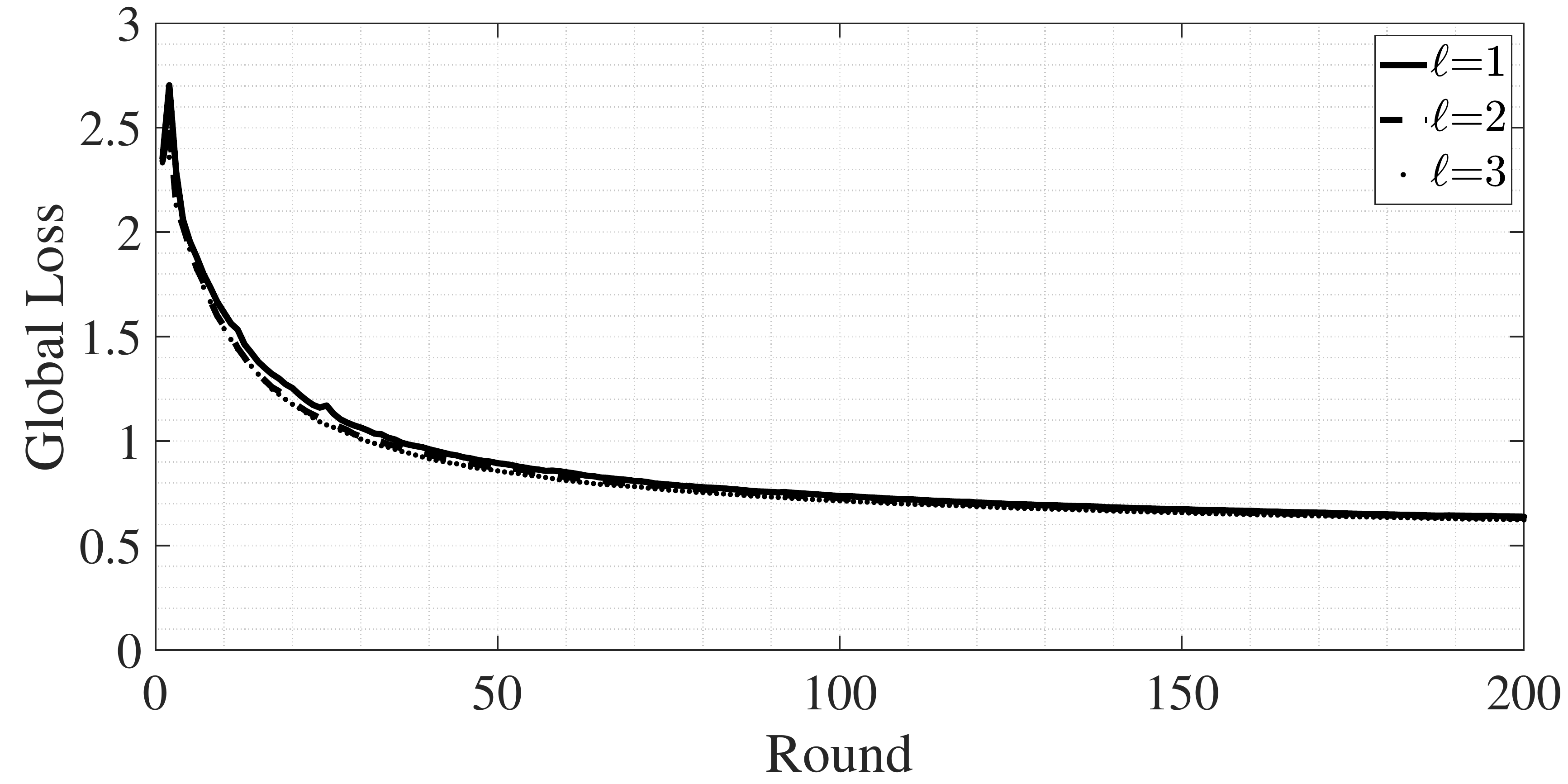}}
	\subfigure[Accuracy.]{\includegraphics[width=0.45\textwidth]{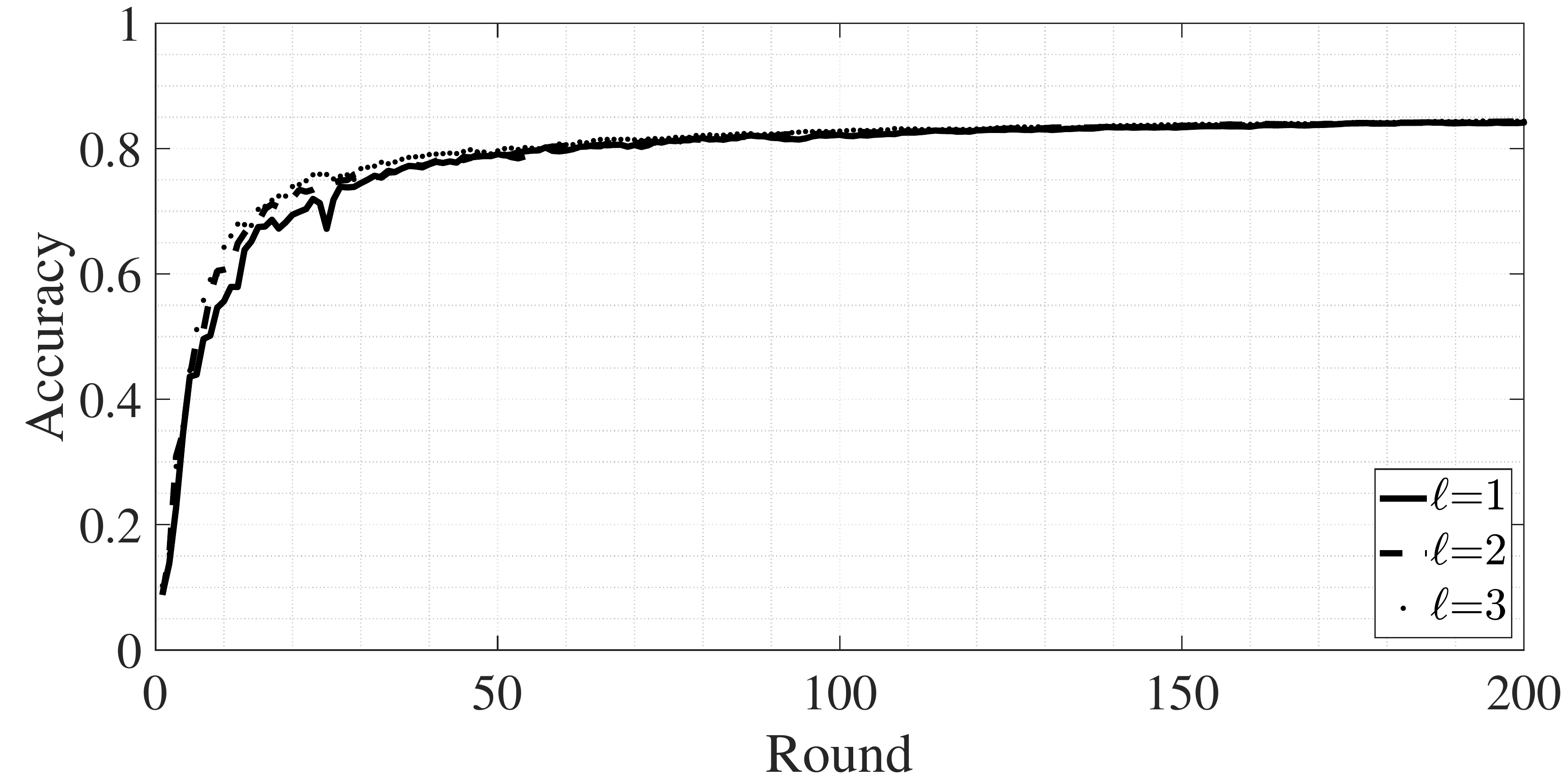}}
	\caption{Performance of the proposed FL with Scheme I sampling over MNIST dataset. In (a) and (b), $\ell=1$. In (c) and (d), $E=2$.}
	\label{fig:SchemeI_MNIST}
\end{figure}

\begin{figure}
	\centering     
	\subfigure[Global loss.]{\includegraphics[width=0.45\textwidth]{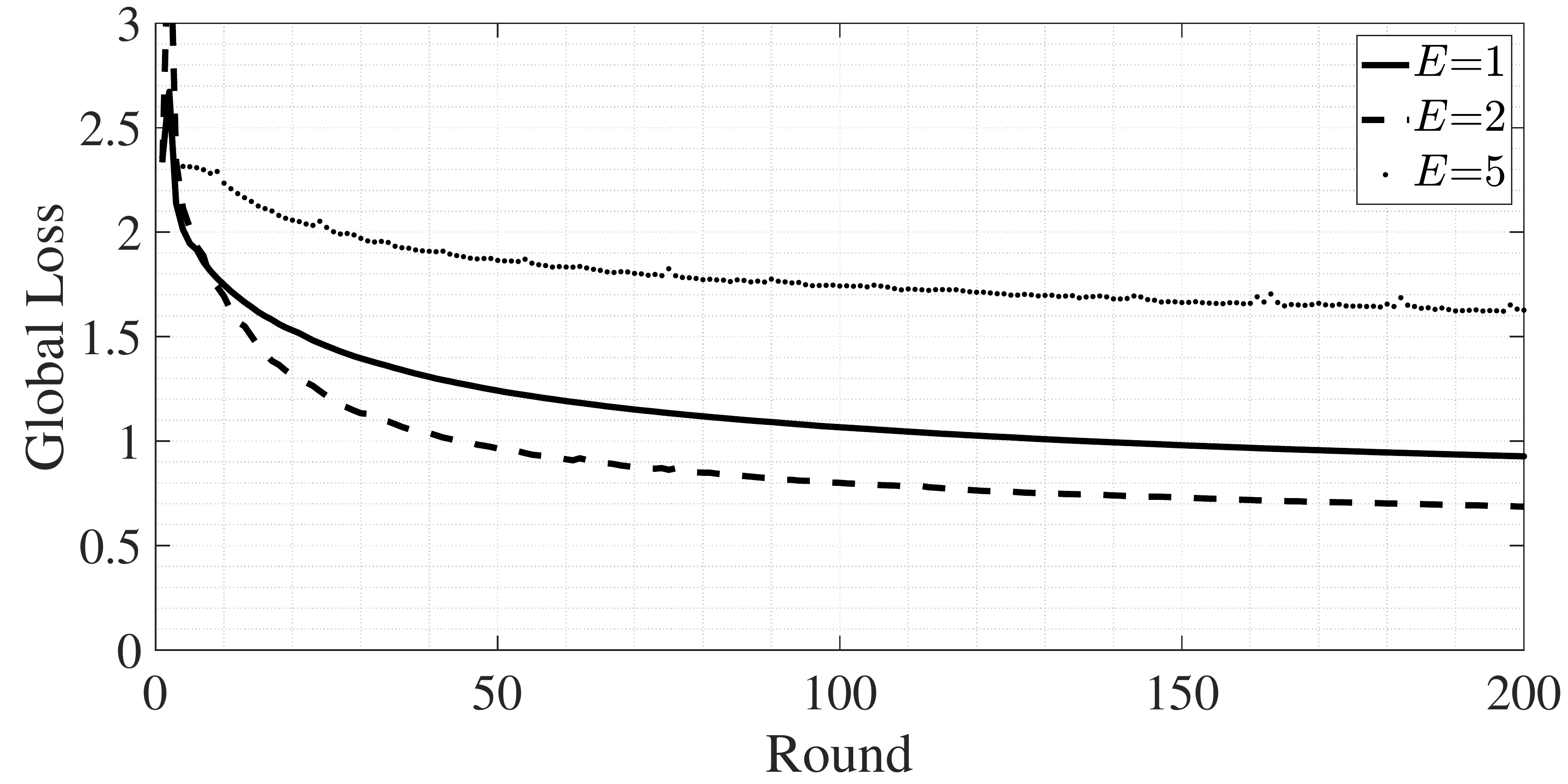}}
	\subfigure[Accuracy.]{\includegraphics[width=0.45\textwidth]{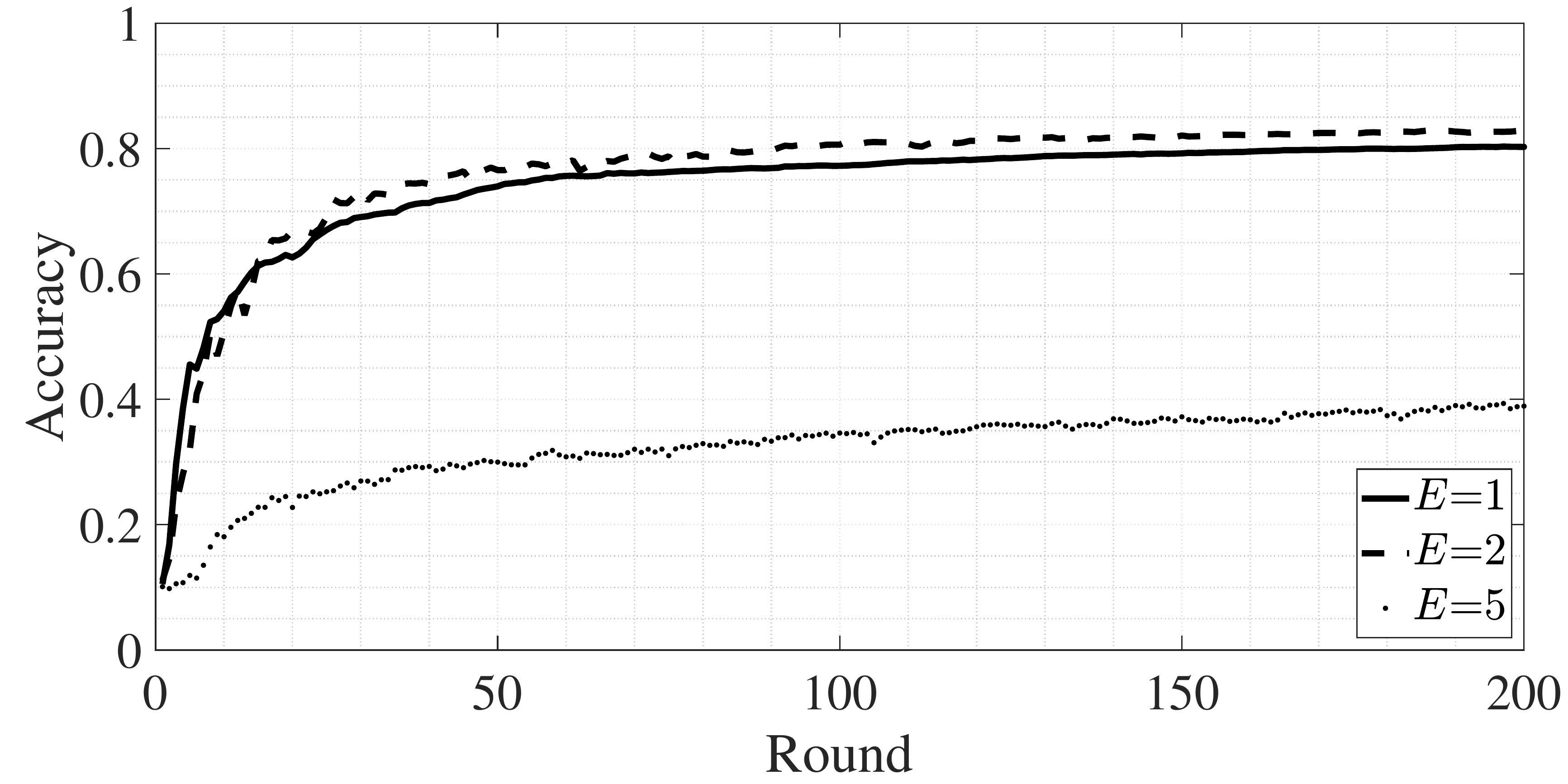}}
	\subfigure[Global loss.]{\includegraphics[width=0.45\textwidth]{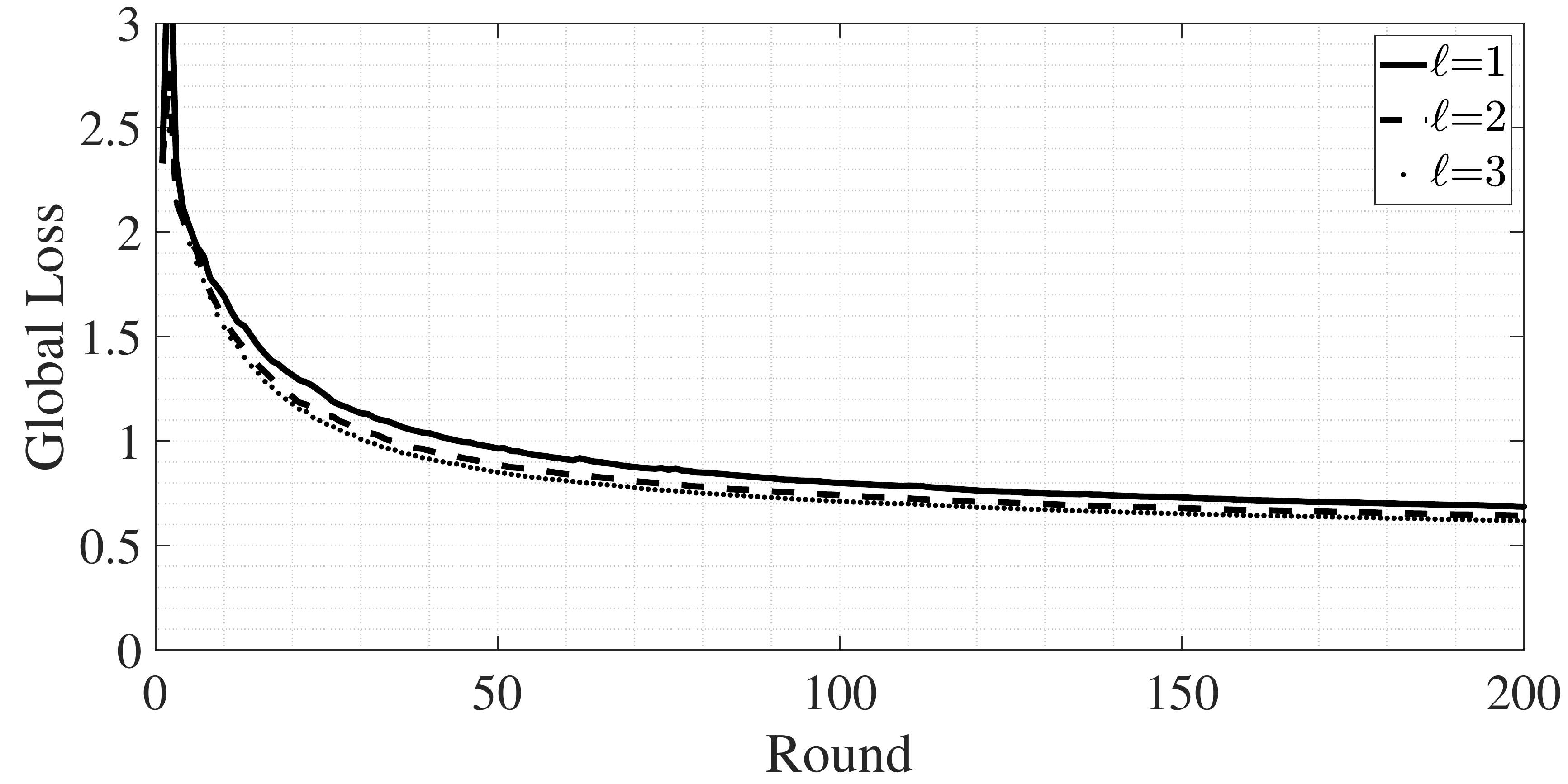}}
	\subfigure[Accuracy.]{\includegraphics[width=0.45\textwidth]{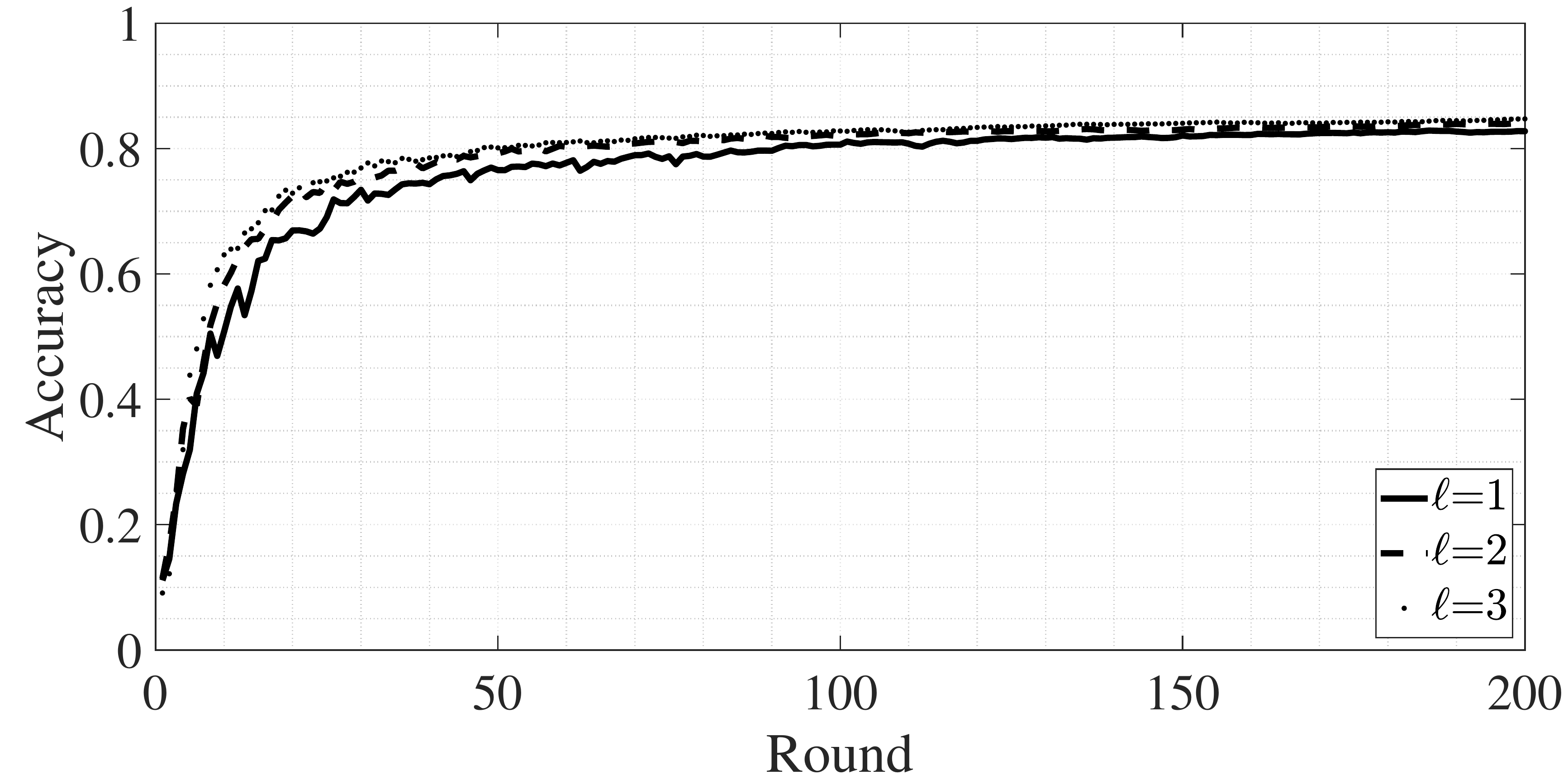}}
	\caption{Performance of the proposed FL with Scheme II sampling with $\hat{q}_k=\frac{1}{N}$, for all $k$, over MNIST dataset. In (a) and (b), $\ell=1$. In (c) and (d), $E=2$.}
	\label{fig:SchemeII_MNIST}
\end{figure}

\begin{figure}
	\centering     
	\subfigure[Global loss.]{\includegraphics[width=0.45\textwidth]{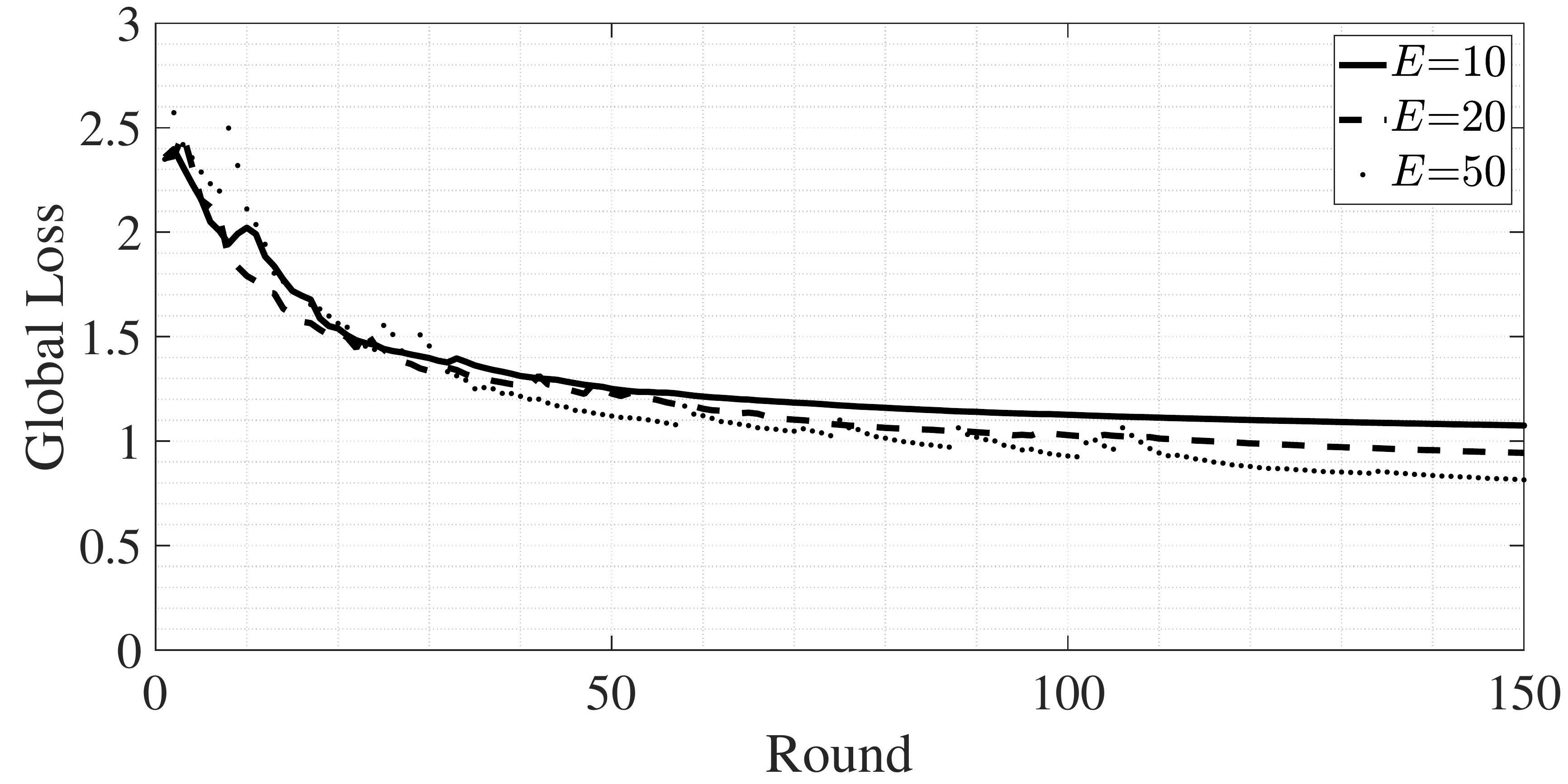}}
	\subfigure[Accuracy.]{\includegraphics[width=0.45\textwidth]{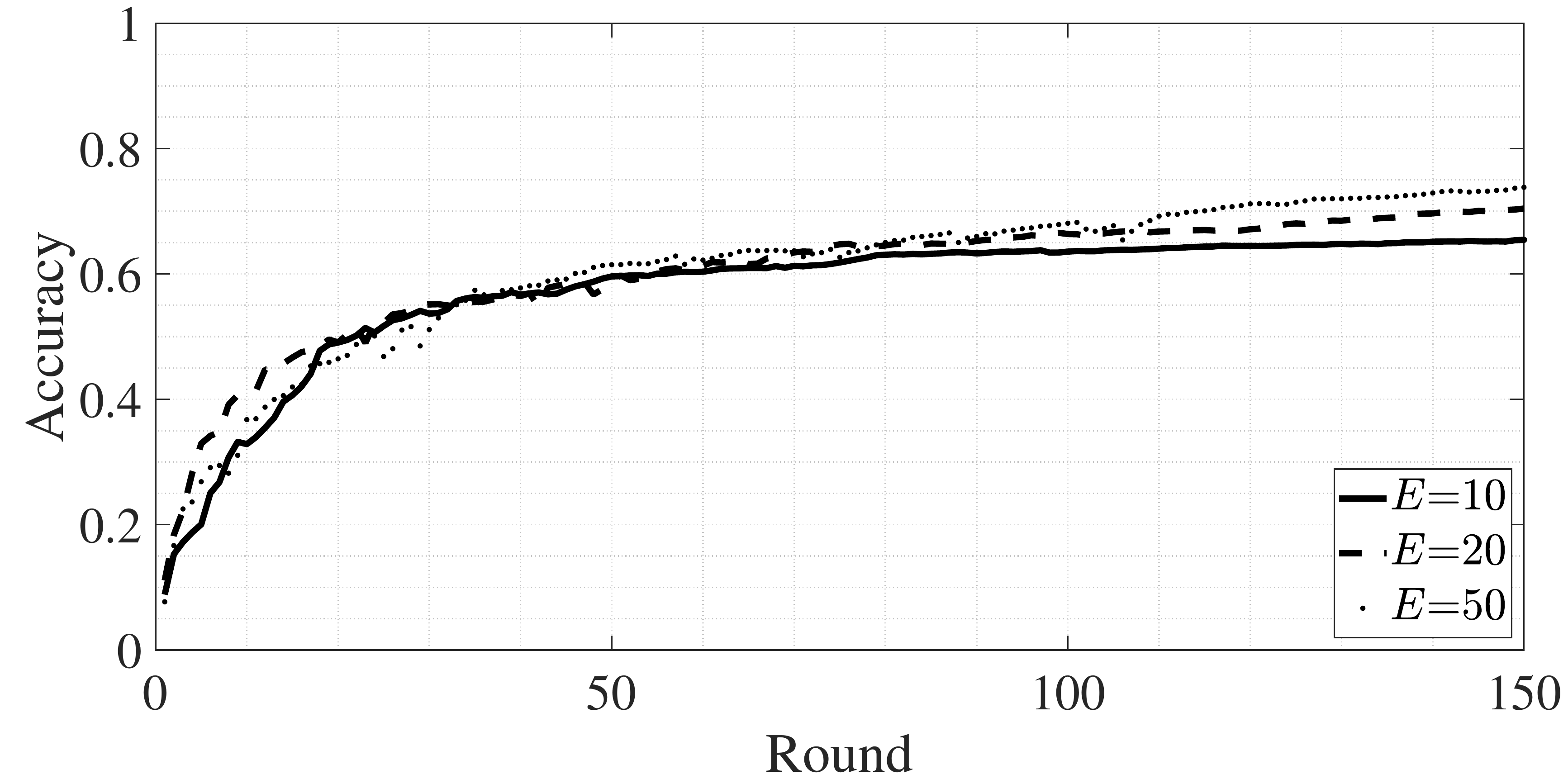}}
	\subfigure[Global loss.]{\includegraphics[width=0.45\textwidth]{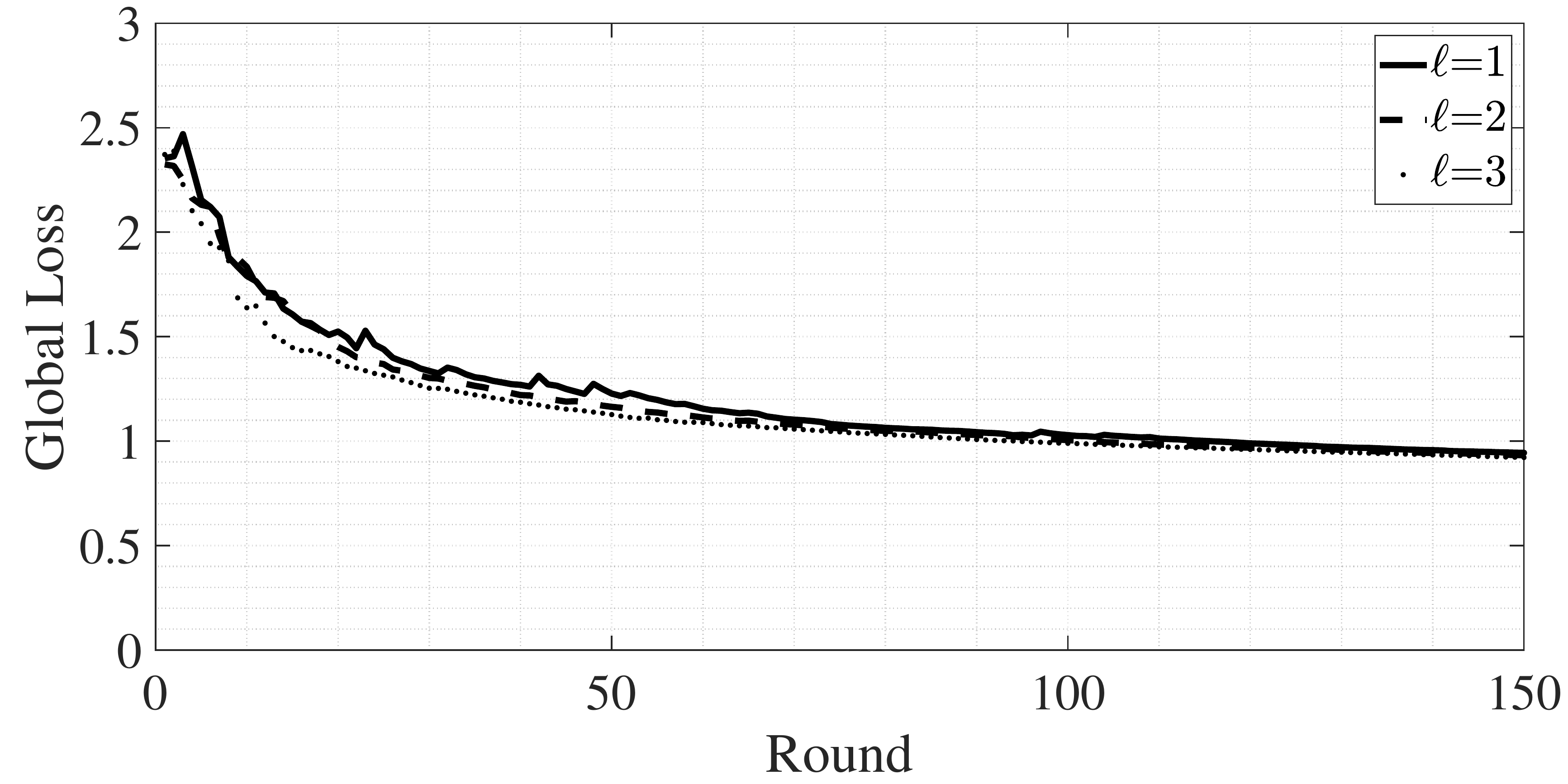}}
	\subfigure[Accuracy.]{\includegraphics[width=0.45\textwidth]{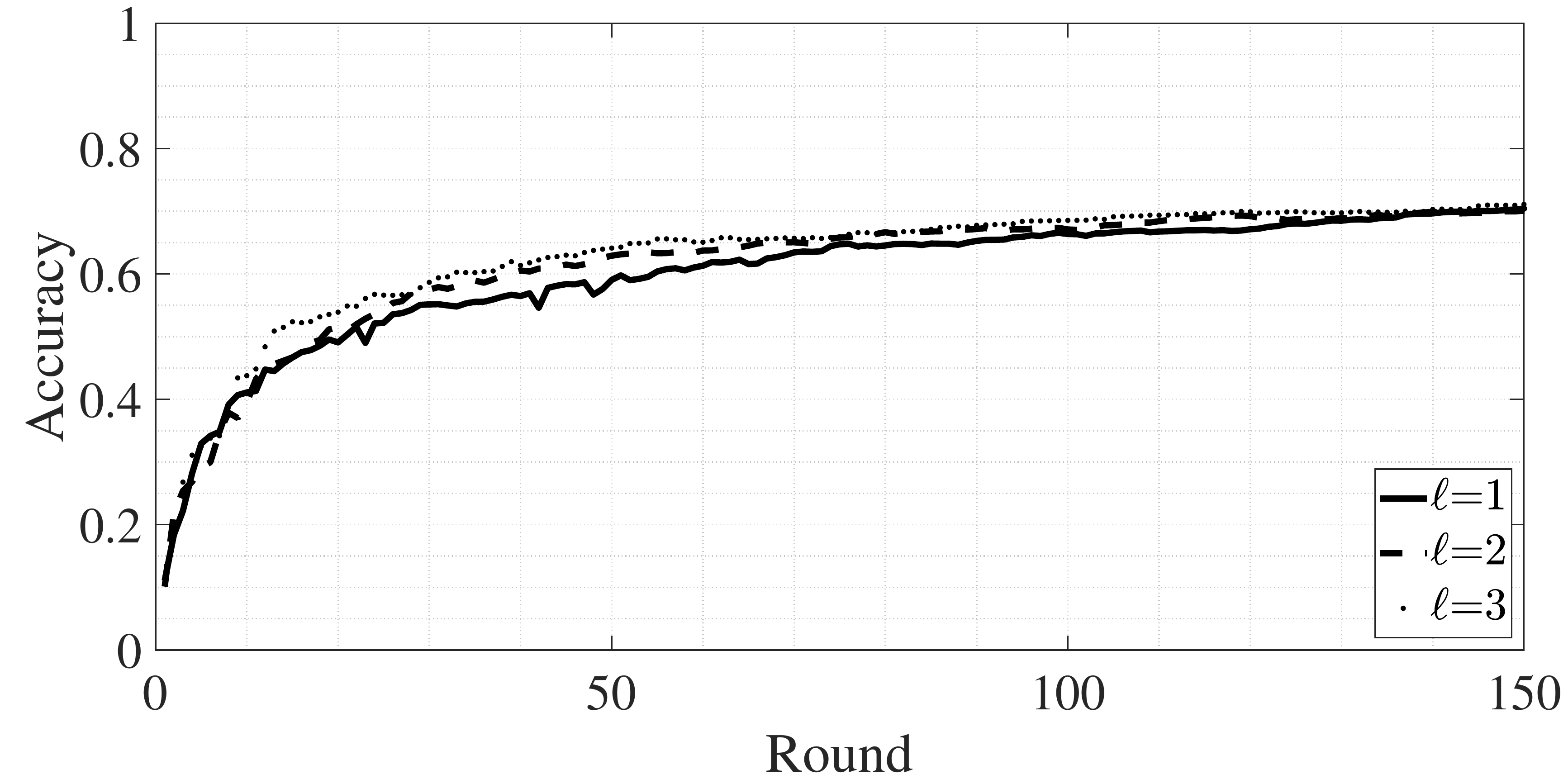}}
	\caption{Performance of the proposed FL with Scheme I sampling over the synthetic dataset. In (a) and (b), $\ell=1$. In (c) and (d), $E=20$.}
	\label{fig:SchemeI_SYNTHETIC}
\end{figure}

\begin{figure}
	\centering     
	\subfigure[Global loss.]{\includegraphics[width=0.45\textwidth]{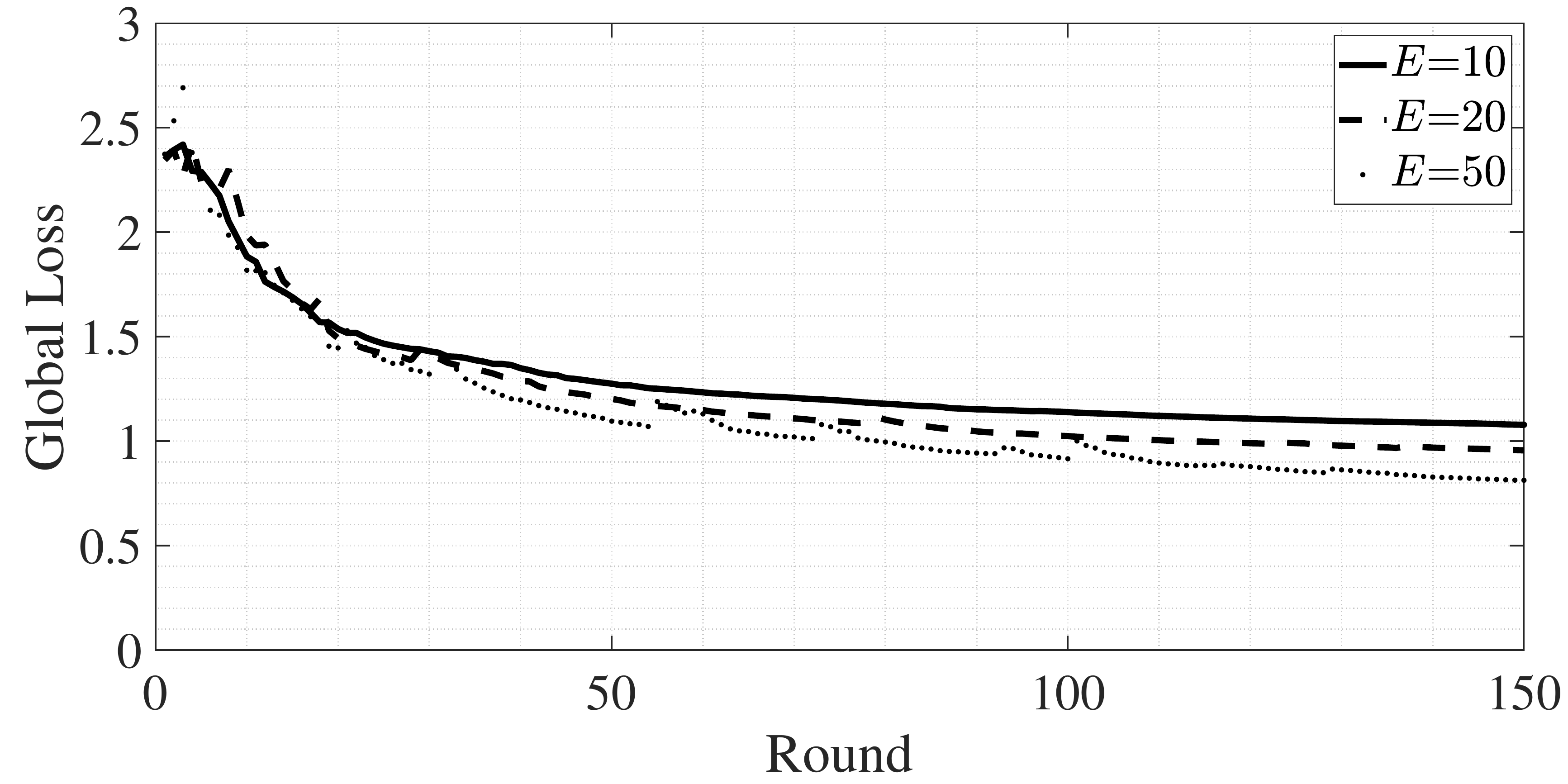}}
	\subfigure[Accuracy.]{\includegraphics[width=0.45\textwidth]{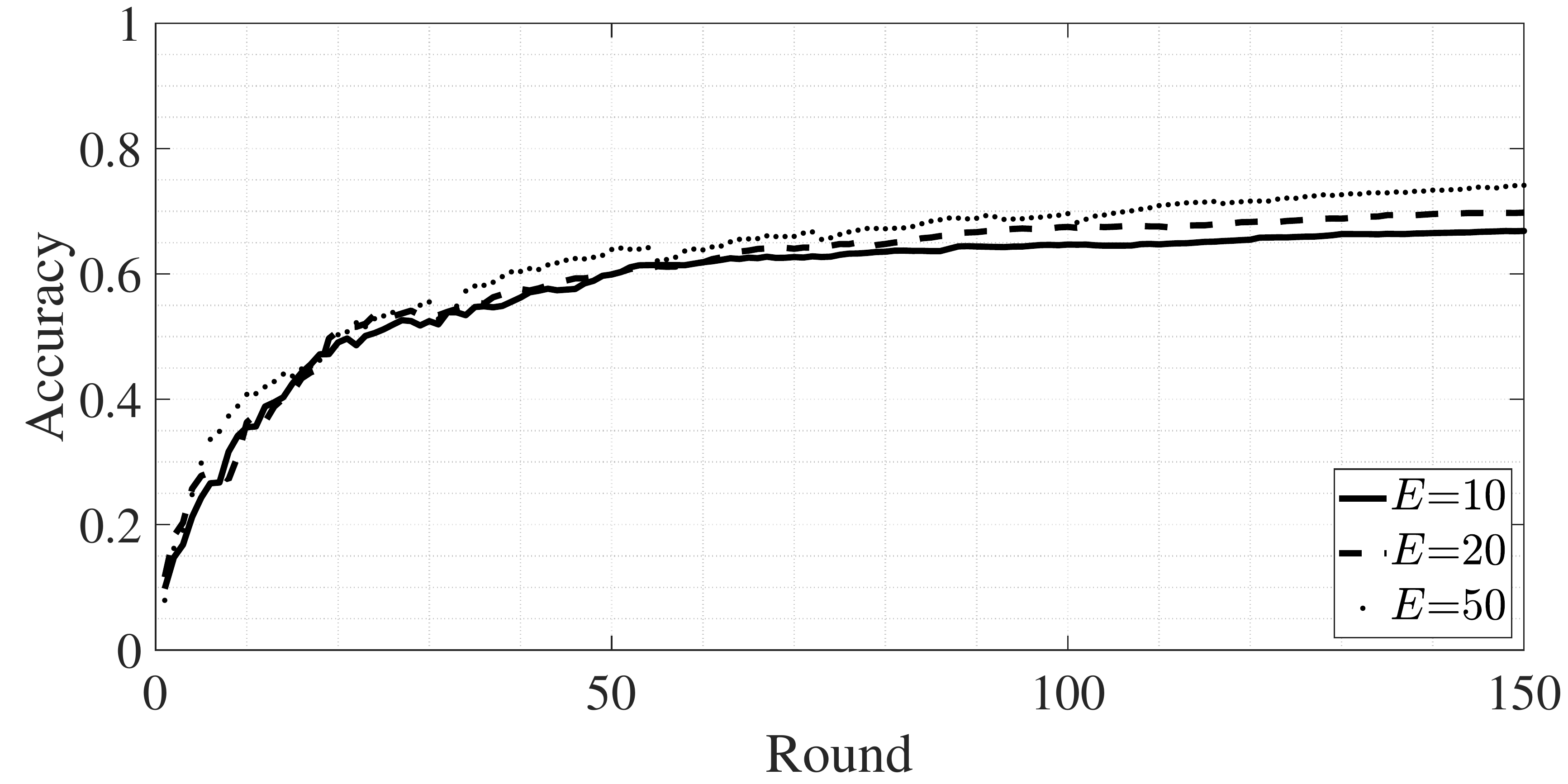}}
	\subfigure[Global loss.]{\includegraphics[width=0.45\textwidth]{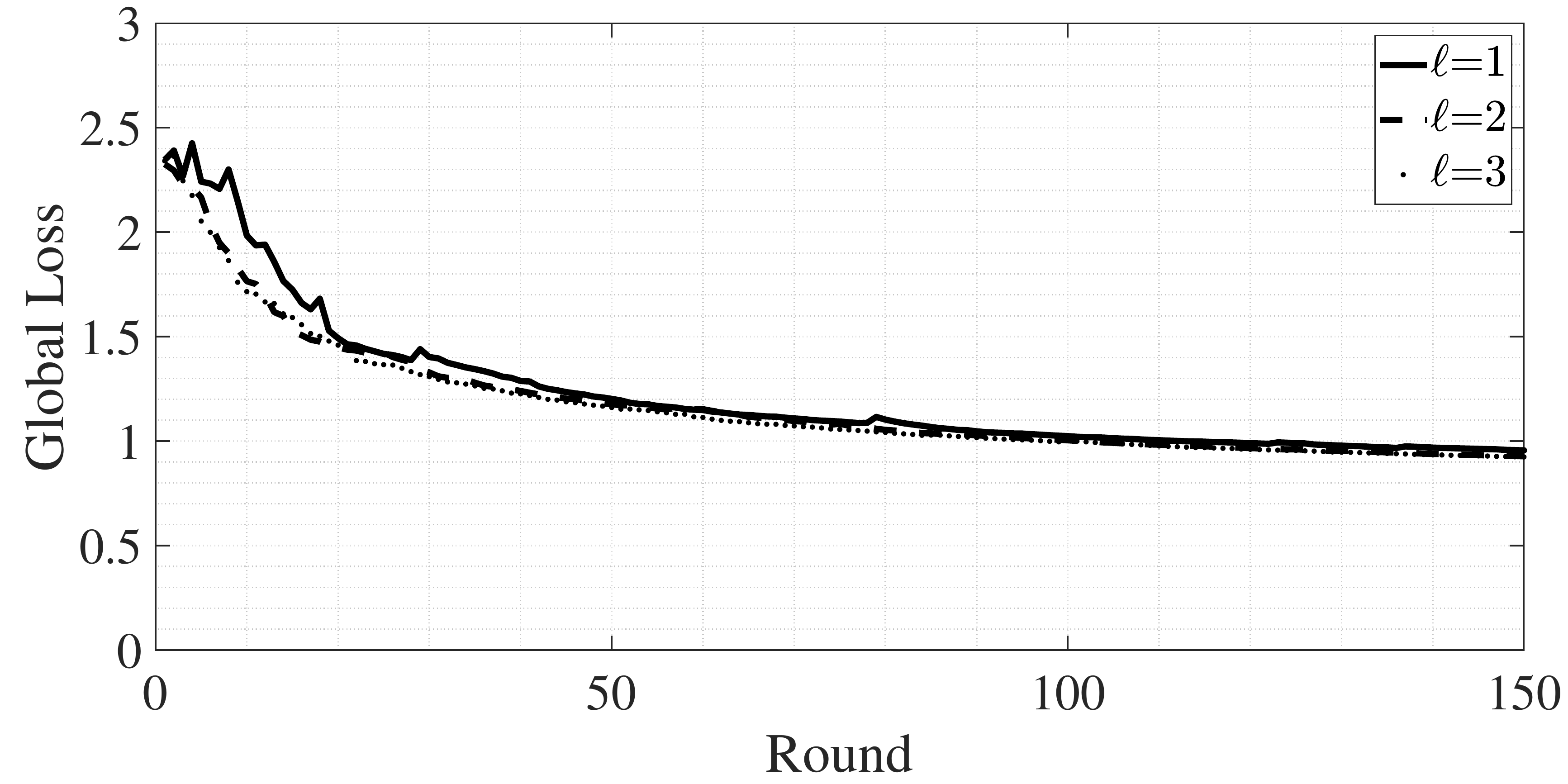}}
	\subfigure[Accuracy.]{\includegraphics[width=0.45\textwidth]{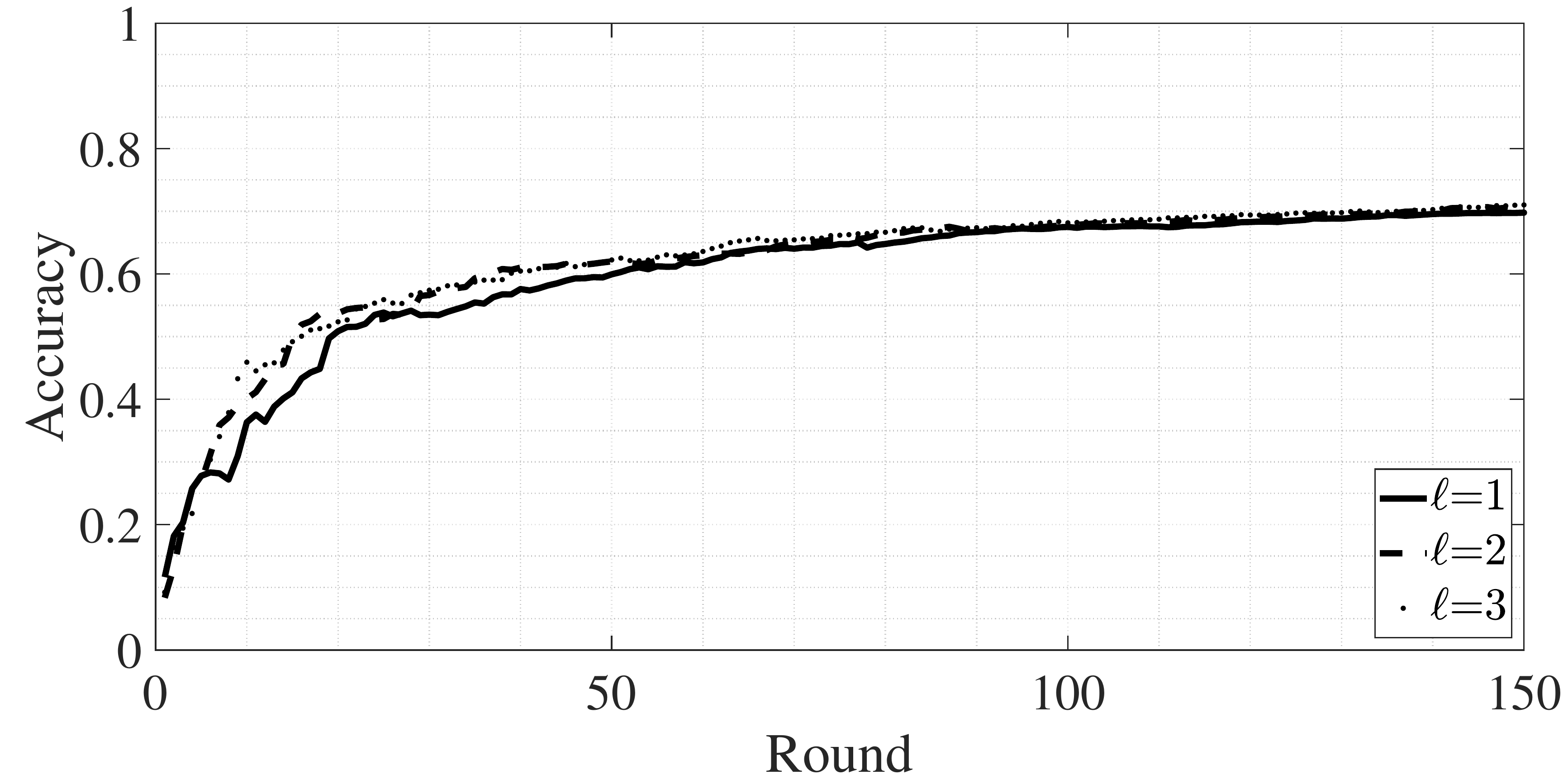}}
	\caption{Performance of the proposed FL with Scheme II sampling with $\hat{q}_k=\frac{1}{N}$, for all $k$, over the synthetic dataset. In (a) and (b), $\ell=1$. In (c) and (d), $E=20$.}
	\label{fig:SchemeII_SYNTHETIC}
\end{figure}

We provide more simulation results in \figref{fig:SchemeI_MNIST}, \ref{fig:SchemeII_MNIST}, \ref{fig:SchemeI_SYNTHETIC}, and \ref{fig:SchemeII_SYNTHETIC} where we study the effects of $K$, $\ell$, and $E$ on the performance of our algorithm. For {\em Scheme I} sampling, we show the results in \figref{fig:SchemeI_MNIST} and \ref{fig:SchemeI_SYNTHETIC}. For {\em Scheme II} sampling with $\hat{q}_k=\frac{1}{N}$, we show the results in \figref{fig:SchemeII_MNIST} and \ref{fig:SchemeII_SYNTHETIC}. Note that, when $\hat{q}_k=\frac{1}{N}$ in {\em Scheme II}, both schemes select users uniformly; however, {\em Scheme I} selects without replacement, while replacement is allowed in {\em Scheme II}. We see that {\em Scheme I} performs better than {\em Scheme II} with uniform selection. This can be also understood from \textbf{Theorem 1}. Finally, it is worth mentioning that, with appropriate tuning of the hyper-parameters ($K$, $\ell$, and $E$), we can reach the benchmarked performance levels.

\section{Conclusion}
Federated learning algorithms cannot be employed in the real world scenarios unless they consider the scarcity of radio resources and unreliability of wireless transmissions. In this regard, in this paper, we have proposed a federated learning algorithm that is tailored for unreliable and resource-constrained wireless networks, where success probability varies across different devices. Our proposed federated learning algorithm incorporates the success probability in the averaging step. Thus, as the first step towards designing the FL algorithm, we have used stochastic geometry tools to calculate the success probability. We have studied the convergence of the proposed algorithm. Specifically, we have proven that the algorithm converges with rate $\mathcal{O}(\frac{1}{T})$ for strongly convex and smooth problems on non-i.i.d. data. The effects of computation, communication, and scheduling on the convergence rate of the algorithm have also been investigated. Finally, we have verified our algorithm through experimenting on real and synthetic datasets.      

\section*{Appendix A: Proof of Lemma~1}
\renewcommand{\theequation}{A.\arabic{equation}}
\setcounter{equation}{0}
For device $k$ at distance $r_k$ from its associated BS, the success probability is derived as follows
\allowdisplaybreaks{
\begin{IEEEeqnarray}{rCl}
	U_k 
	&=& 1-\mathbb{E}\left[ \mathbf{1} \left( \max\left\{\text{SINR}_{k}(1),\text{SINR}_{k}(2),...,\text{SINR}_{k}(\ell) \right\}<\theta \right) \mid k\in \mathcal{S}_t \right] 
	\nonumber \\
	&=& 1-\mathbb{E}_{\Phi_{\rm I}}\left[ \prod_{i=1}^{\ell} \mathbb{E}\left[ \mathbf{1}\left( \text{SINR}_{k}(i) <\theta \right) \right] \mid k\in \mathcal{S}_t \right]
	\nonumber \\
	&=& 1-\mathbb{E}_{\Phi_{\rm I}}\left[ \prod_{i=1}^{\ell} \mathbb{E}\left[ \mathbf{1}\left( \frac{h_{k}(i)r_k^{-\alpha}}{I(i) +\sigma^2} < \theta \right) \right] \right]
	\nonumber \\
	&\stackrel{\text{(a)}}{=}& 1-\mathbb{E}_{\Phi_{\rm I}}\left[ 
	\left( 1-e^{-\theta\sigma^2r_k^{\alpha}} \prod_{x\in\Phi_{\rm I}} \frac{1}{1+\theta r_k^{\alpha}\|x\|^{-\alpha}} \right)^\ell 
	\right]
	\label{eq:A-mid_step} \\
	&\stackrel{\text{(b)}}{=}& \sum_{i=1}^{\ell} \binom{\ell}{i} (-1)^{i+1} e^{-i\theta\sigma^2r_k^{\alpha}} \,\mathbb{E}_{\Phi_{\rm I}}\left[ \prod_{x\in\Phi_{\rm I}} \frac{1}{ \left(1+\theta r_k^{\alpha}\|x\|^{-\alpha}\right)^i }
	\right]
	\nonumber \\
	&\stackrel{\text{(c)}}{\approx}& \sum_{i=1}^{\ell} \binom{\ell}{i} (-1)^{i+1} 
	\exp\left\{ -i\theta\sigma^2r_k^{\alpha}-\int_{\mathbb{R}^2} \left(1-\frac{1}{ \left(1+\theta r_k^{\alpha}\|x\|^{-\alpha}\right)^i } \right) \lambda (1-e^{-12/5\lambda\pi\|x\|^2}) {\rm d}x \right\}
	\nonumber \\
	&\stackrel{\text{(d)}}{=}& \sum_{i=1}^{\ell} \binom{\ell}{i} (-1)^{i+1} 
	\exp\left\{-i\theta\sigma^2r_k^{\alpha}
	-2\pi\lambda \int_{0}^{\infty} \left(1-\frac{1}{ \left(1+\theta r_k^{\alpha}x^{-\alpha}\right)^i } \right) (1-e^{-12/5\lambda\pi x^2}) x {\rm d}x \right\}, \nonumber 
\end{IEEEeqnarray}}
where (a) is obtained by averaging with respect to the fading. (b) follows from using the binomial expansion. In reality, $\Phi_{\rm I}$ is a subset of the set of the participating devices in the federated learning. However, since $N$ is large in real-world applications, the BS observes a completely different realization at the time of aggregation after each sampling step. Thus we can approximate  $\Phi_{\rm I}$ by a non-homogeneous PPP with intensity function $\lambda(x)=\lambda\left(1-e^{-12/5\lambda\pi\|x\|^2}\right)$ \cite{wang2017meta}. In (c), we use  the probability generating functional (PGFL) for the point process $\Phi_{\rm I}$. Finally, (d) is obtained by using the polar domain representation. 

\section*{Appendix B: Proof of Theorem~1\footnote{We follow the same procedure as in \cite{li2019convergence}. Therefore, we mainly focus on the parts that are different.  }}
\renewcommand{\theequation}{B.\arabic{equation}}
\setcounter{equation}{0}
When we choose the learning rate as $\eta_t=\frac{\beta}{\mu(\gamma+t)}$ with $\beta>0$ and $\gamma=\max\left\{4\beta\frac{L}{\mu},E\right\}$, it fulfills the following properties: i) $\eta_t$ is decreasing with respect to $t$, ii) $\eta_t \le \frac{1}{2L}$, $\forall t$, and iii) $\eta_t \le 2\eta_{t+E}$, $\forall t$. These properties help us with the convergence analysis of the proposed federated learning algorithm. 

In the following, we provide three lemmas which help us in proving the theorem.
\begin{Lemma} Averaging step in the proposed algorithm is unbiased, i.e.
	\begin{IEEEeqnarray}{rCl}
	\mathbb{E}\left[ \bar{w}_t \right]=\bar{v}_t, \qquad t\in\mathcal{I}_E, \nonumber 
	\end{IEEEeqnarray}
	where the expectation is over sampling and success event.
\end{Lemma}
\begin{IEEEproof}
	When $t\in\mathcal{I}_E$, $\bar{w}_t=w_t$; hence, from \eqref{eq:Averaging}, we have
	\begin{IEEEeqnarray}{rCl}
		\IEEEeqnarraymulticol{3}{l}{ \mathbb{E}\left[ \bar{w}_t \right] = 
		\mathbb{E}\left[   
		w_{t-E} + \sum_{k=1}^{N} \sum_{m=1}^{M} \frac{p_k}{q_kU_k} \mathbf{1}\left(k\in \mathcal{S}_{t-E }(m), \text{SINR}_{k,m}>\theta\right) (v_t^k-w_{t-E})
		\right] }
		\nonumber \\
		&\stackrel{\text{(a)}}{=}& w_{t-E} + \sum_{k=1}^{N} \frac{p_k}{q_kU_k} \mathbb{E}\left[ \sum_{m=1}^{M} \mathbf{1}\left(k\in \mathcal{S}_{t-E}(m)\right) 
		\mathbb{E}\left[ \mathbf{1}\left(\text{SINR}_{k,m}>\theta\right) \mid k\in \mathcal{S}_{t-E}(m)\right] \right] (v_t^k-w_{t-E}) 
		\nonumber \\
		&=& w_{t-E} + \sum_{k=1}^{N} \frac{p_k}{q_kU_k} \mathbb{E}\left[ \sum_{m=1}^{M} \mathbf{1}\left(k\in \mathcal{S}_{t-E}(m)\right) 
		U_k \right] (v_t^k-w_{t-E}) \nonumber \\
		&=& w_{t-E} + \sum_{k=1}^{N} \frac{p_k}{q_kU_k}  q_k U_k  (v_t^k-w_{t-E}) \nonumber \\
		&=& \sum_{k=1}^{N} p_k v_t^k = \bar{v}_t, \nonumber
	\end{IEEEeqnarray}
	where, in (a), the inner expectation is over the success event, and the outer expectation is with respect to the sampling.
\end{IEEEproof}

\begin{Lemma} When $\eta_t$ satisfies the aforementioned properties,   
	\begin{IEEEeqnarray}{rCl}
		\mathbb{E}\left[ \| \bar{w}_{t+1}-\bar{v}_{t+1} \|^2 \right] \le 4 \eta_t^2 E^2 G^2 B, \qquad
		{t+1}\in\mathcal{I}_E, \nonumber 
	\end{IEEEeqnarray}
	where $B=\sum_{k=1}^N p_k \left( \frac{1}{q_k U_k}-1 \right)$ for sampling Scheme I, and $B=\sum_{k=1}^N  p_k\left( \frac{1}{q_k U_k}-\frac{1}{M} \right)$ for sampling {\em Scheme II}. The expectation is over sampling and success event.
\end{Lemma}
\begin{IEEEproof} For brevity, in this proof, we denote the set of scheduled devices at time $t+1-E$ by $\mathcal{S}$ rather than $\mathcal{S}_{t+1-E}$.
	\allowdisplaybreaks{
	\begin{IEEEeqnarray}{rCl}
		\IEEEeqnarraymulticol{3}{l}{ \mathbb{E}\left[ \| \bar{w}_{t+1}-\bar{v}_{t+1} \|^2 \right] } 
		\nonumber \\
		&=& \mathbb{E}\left[ \left\| w_{t+1-E} + \sum_{k=1}^{N} \sum_{m=1}^{M} \frac{p_k}{q_kU_k} \mathbf{1}\left(k\in \mathcal{S}(m), \text{SINR}_{k,m}>\theta\right) (v_{t+1}^k-w_{t+1-E}) - \sum_{k=1}^{N} p_k v_{t+1}^k \right\|^2 \right] 
		\nonumber \\
		&=& \mathbb{E}\left[ \left\| \sum_{k=1}^{N} \sum_{m=1}^{M} \frac{p_k}{q_kU_k} \mathbf{1}\left(k\in \mathcal{S}(m), \text{SINR}_{k,m}>\theta\right) (v_{t+1}^k-w_{t+1-E}) - \sum_{k=1}^{N} p_k (v_{t+1}^k-w_{t+1-E}) \right\|^2 \right] 
		\nonumber \\
		&=& \mathbb{E}\left[ \left\| \sum_{k=1}^{N} p_k \frac{\sum_{m=1}^{M} \mathbf{1}\left(k\in \mathcal{S}(m), \text{SINR}_{k,m}>\theta\right)-q_kU_k}{q_kU_k}
		(v_{t+1}^k-w_{t+1-E}) \right\|^2 \right] 
		\nonumber \\
		&\stackrel{\text{(a)}}{\le}& \mathbb{E}\left[ \sum_{k=1}^{N} p_k \left\| \frac{\sum_{m=1}^{M} \mathbf{1}\left(k\in \mathcal{S}(m), \text{SINR}_{k,m}>\theta\right)-q_kU_k}{q_kU_k}
		(v_{t+1}^k-w_{t+1-E}) \right\|^2 \right] 
		\nonumber \\
		&\stackrel{\text{(b)}}{=}& \sum_{k=1}^{N} p_k 
		\frac{ \mathbb{E}\left[ \left( \sum_{m=1}^{M} \mathbf{1}\left(k\in \mathcal{S}(m), \text{SINR}_{k,m}>\theta\right) \right)^2 \right]-q_k^2U_k^2}{q_k^2U_k^2}
		\mathbb{E}\left[ \left\| v_{t+1}^k-w_{t+1-E} \right\|^2 \right], \label{eq:B1}
	\end{IEEEeqnarray}}
	where (a) is obtained from convexity of $\|.\|^2$, and (b) follows from 
	\begin{IEEEeqnarray}{rCl}
		\mathbb{E}\left[ \sum_{m=1}^{M} \mathbf{1}\left(k\in \mathcal{S}(m), \text{SINR}_{k,m}>\theta\right) \right]=q_kU_k,
		\nonumber
	\end{IEEEeqnarray}	
	where the expectation is with respect to the sampling and success event. In the following, we first calculate $\mathbb{E}\left[ \left\| v_{t+1}^k-w_{t+1-E} \right\|^2 \right]$, and then calculate $\mathbb{E}\left[ \left( \sum_{m=1}^{M} \mathbf{1}\left(k\in \mathcal{S}(m), \text{SINR}_{k,m}>\theta\right) \right)^2 \right]$ for each sampling scheme separately.
	\begin{IEEEeqnarray}{rCl}
		\mathbb{E}\left[ \left\| v_{t+1}^k-w_{t+1-E} \right\|^2 \right] &=& \mathbb{E}\left[ \left\| \sum_{i=t+1-E}^t \eta_i \nabla F_k (w_i^k;\xi_i^k) \right\|^2 \right] \nonumber \\
		&\stackrel{\text{(a)}}{\le}& \mathbb{E}\left[ E \sum_{i=t+1-E}^t \left\| \eta_i \nabla F_k (w_i^k;\xi_i^k) \right\|^2 \right]
		\nonumber \\
		&\stackrel{\text{(b)}}{\le}& \mathbb{E}\left[ \eta_{t+1-E}^2 E \sum_{i=t+1-E}^t \left\| \nabla F_k (w_i^k;\xi_i^k) \right\|^2 \right]
		\nonumber \\
		&\stackrel{\text{(c)}}{\le}& \eta_{t+1-E}^2 E^2 G^2 \stackrel{\text{(d)}}{\le} 4 \eta_{t+1}^2  E^2 G^2 \le 4 \eta_{t}^2  E^2 G^2, \label{eq:B2}
	\end{IEEEeqnarray}
	where (a) is obtained by using Cauchy-Schwarz inequality. (b) follows from the property that $\eta_t$ is decreasing with respect to $t$. In (c), we have used \textbf{Assumption 4}. (d) follows from $\eta_t \le 2 \eta_{t+E}$.
	
	In {\em Scheme I}, at the aggregation step, each device can use at most one resource block. Thus,
	\begin{IEEEeqnarray}{rCl}
		\mathbb{E}\left[ \left( \sum_{m=1}^{M} \mathbf{1}\left(k\in \mathcal{S}(m), \text{SINR}_{k,m}>\theta\right) \right)^2 \right] = 
		\mathbb{E}\left[ \sum_{m=1}^{M} \mathbf{1}\left(k\in \mathcal{S}(m), \text{SINR}_{k,m}>\theta\right) \right]=q_kU_k. \nonumber \\ 
		\label{eq:B3}
	\end{IEEEeqnarray} 
	However, in the second scheme, the BS may allocate more than one resource block to a device at a sampling time. $\mathbf{1}\left(k\in \mathcal{S}(m), \text{SINR}_{k,m}>\theta\right)$ is a Bernoulli random variable which takes value one with probability $\hat{q}_kU_k$. For {\em Scheme II}, sampling and success event over each resource block are i.i.d.; therefore,  $\sum_{m=1}^{M} \mathbf{1}\left(k\in \mathcal{S}(m), \text{SINR}_{k,m}>\theta\right)$ is distributed according to a binomial distribution with parameters $M$ and $\hat{q}_kU_k$, and we have
	\begin{IEEEeqnarray}{rCl}
		\mathbb{E}\left[ \left( \sum_{m=1}^{M} \mathbf{1}\left(k\in \mathcal{S}(m), \text{SINR}_{k,m}>\theta\right) \right)^2 \right] &=&  
		q_kU_k \left( 1-\frac{1}{M}q_kU_k \right) + q_k^2U_k^2
		\nonumber \\
		&=& q_kU_k+\left( 1-\frac{1}{M} \right) q_k^2 U_k^2.
		\label{eq:B4}
	\end{IEEEeqnarray} 
\end{IEEEproof}

\begin{Lemma} When $\eta_t$ satisfies the aforementioned properties, for any $t$, we have
	\begin{IEEEeqnarray}{rCl}
		\mathbb{E}\left[ \| \bar{v}_{t+1}-w^* \|^2 \right] \le
		(1-\mu\eta_t) \mathbb{E}\left[ \| \bar{w}_{t}-w^* \|^2 \right] + \eta_t^2 \left( \sum_{k=1}^{N} p_k^2\sigma_k^2 + 6L\Gamma + 8(E-1)^2G^2 \right). \nonumber
	\end{IEEEeqnarray}
\end{Lemma}
\begin{IEEEproof}
	See \textbf{Appendix A} in \cite{li2019convergence}.
\end{IEEEproof}

Note that \textbf{Lemma 2} and \textbf{Lemma 3} only consider $t\in\mathcal{I}_E$. When, $t\not\in\mathcal{I}_E$, $\bar{w}_t=\bar{v}_t$ according to \eqref{eq:Averaging}.

Based on the above discussion, at any $t$, we have
\begin{IEEEeqnarray}{rCl}
	\IEEEeqnarraymulticol{3}{l}{
	\mathbb{E}\left[ \left\| \bar{w}_{t+1}-w^* \right\|^2 \right] = \mathbb{E}\left[ \left\| \bar{w}_{t+1}-\bar{v}_{t+1}+\bar{v}_{t+1}-w^* \right\|^2 \right] }
	\nonumber \\
	&\stackrel{\text{(a)}}{=}& \mathbb{E}\left[ \left\| \bar{w}_{t+1}-\bar{v}_{t+1} \right\|^2 \right] 
	+ \mathbb{E}\left[ \left\| \bar{v}_{t+1}-w^* \right\|^2 \right] 
	+ 2 \mathbb{E}\left[  \left( \bar{w}_{t+1}-\bar{v}_{t+1} \right)^T \left( \bar{v}_{t+1}-w^* \right) \right] 
	\nonumber \\
	&\stackrel{\text{(b)}}{\le}&
	(1-\mu\eta_t) \mathbb{E}\left[ \| \bar{w}_{t}-w^* \|^2 \right] + \eta_t^2 \left( \sum_{k=1}^{N} p_k^2\sigma_k^2 + 6L\Gamma + 8(E-1)^2G^2 + 4E^2G^2B \right),
	\label{eq:B5}
\end{IEEEeqnarray}
where the last term in (a) is zero based on \textbf{Lemma 2} and the fact that $\bar{w}_t=\bar{v}_t$ when $t\not\in\mathcal{I}_E$. In (b), we have used \textbf{Lemma 3} and \textbf{Lemma 4}. 

For brevity, we define $C=\sum_{k=1}^{N} p_k^2\sigma_k^2 + 6L\Gamma + 8(E-1)^2G^2 + 4E^2G^2B$  and $\Delta_t = \| \bar{w}_t-w^* \|^2$. In the following, we find $v$ such that $\mathbb{E}\left[ \Delta_t \right] \le \frac{v}{\gamma+t}$ at any $t$ after initializing with $\Delta_0$. This is satisfied at time $t=0$ when $v \ge \gamma \Delta_0$. Moreover, when $\beta>1$ and $v\ge\frac{\beta^2 C}{\mu^2(\beta-1)}$, $\mathbb{E}\left[ \Delta_{t+1} \right] \le \frac{v}{\gamma+t+1}$ given $\mathbb{E}\left[ \Delta_{t} \right] \le \frac{v}{\gamma+t}$. The proof is as follows:
\begin{IEEEeqnarray}{rCl}
	\mathbb{E}\left[ \Delta_{t+1} \right] &\le& (1-\mu\eta_t) \mathbb{E}\left[ \Delta_{t} \right] + \eta_t^2 C 
	\nonumber \\
	&\le& \left(1-\frac{\beta}{\gamma+t}\right) \frac{v}{\gamma+t} + \frac{\beta^2 C}{\mu^2(\gamma+t)^2} 
	\nonumber \\
	&=& \frac{\gamma+t-1}{(\gamma+t)^2}v + \left[ \frac{\beta^2 C}{\mu^2(\gamma+t)^2} - \frac{\beta-1}{(\gamma+t)^2}v \right] 
	\nonumber \\
	&\stackrel{\text(a)}{\le}& \frac{\gamma+t-1}{(\gamma+t)^2}v \le \frac{v}{\gamma+t+1}, \nonumber
\end{IEEEeqnarray}
where (a) is obtained from $v\ge\frac{\beta^2 C}{\mu^2(\beta-1)}$. Thus, by induction $\mathbb{E}\left[ \Delta_t \right] \le \frac{v}{\gamma+t}$ at any $t$ when $v=\max\left\{\frac{\beta^2 C}{\mu^2(\beta-1)},\gamma \Delta_0\right\}$.

Therefore, when $\eta_t=\frac{2}{\mu(\gamma+t)}$\footnote{We set $\beta=2$.} with $\gamma=\max\{8\frac{L}{\mu},E\}$, we have 
\begin{IEEEeqnarray}{rCl}
	\mathbb{E}\left[ \left\| \bar{w}_{t}-w^* \right\|^2 \right] \le 
	\frac{ \max\left\{ 
		\frac{4}{\mu^2}\left( \sum_{k=1}^{N} p_k^2\sigma_k^2 + 6L\Gamma + 8(E-1)^2G^2 + 4E^2G^2B \right), 
		\gamma \left\| w_{0}-w^* \right\|^2 \right\} }{\gamma+t}, \nonumber \\
	 \label{eq:B6}
\end{IEEEeqnarray}
where $B=\sum_{k=1}^N p_k \left( \frac{1}{q_k U_k}-1 \right)$ for sampling {\em Scheme I}, and $B=\sum_{k=1}^N  p_k\left( \frac{1}{q_k U_k}-\frac{1}{M} \right)$ for sampling Scheme II.

After the averaging step at time $T$ ($T\in\mathcal{I}_E$), from $L$-smoothness of the global objective function $F$, we have 
\begin{IEEEeqnarray}{rCl}
	\IEEEeqnarraymulticol{3}{l}{
	\mathbb{E}\left[ F(w_T)-F^* \right] \le \frac{L}{2} \mathbb{E} \left[ \left\| w_T - w^* \right\|^2 \right] }
	\nonumber \\
	&\stackrel{\text{(a)}}{\le}& \frac{L}{2(\gamma+T)} 	{ \max\left\{ 
	\frac{4}{\mu^2}\left( \sum_{k=1}^{N} p_k^2\sigma_k^2 + 6L\Gamma + 8(E-1)^2G^2 + 4E^2G^2B \right), 
	\gamma \left\| w_{0}-w^* \right\|^2 \right\} }
	\nonumber \\
	&\le& \frac{L/\mu}{\gamma+T} 	\left[ 
		\frac{2}{\mu}\left( \sum_{k=1}^{N} p_k^2\sigma_k^2 + 6L\Gamma + 8(E-1)^2G^2 + 4E^2G^2B \right)+ 
		\frac{\mu \gamma}{2} \left\| w_{0}-w^* \right\|^2 \right], \nonumber
\end{IEEEeqnarray} 
where (a) is obtained from $w_T=\bar{w}_T$ and \eqref{eq:B6}.

\section*{Appendix C}
\renewcommand{\theequation}{C.\arabic{equation}}
\setcounter{equation}{0}
With slight abuse of notation, we define $\hat{F}(w) = \sum_{k=1}^{N} \frac{p_kU_k}{\sum_{k'=1}^N p_{k'} U_{k'}} F_k(w)$, and we prove that \eqref{eq:Compare-Averaging2} solves 
\begin{IEEEeqnarray}{rCl}
	\min_w  \qquad \hat{F}(w)=\sum_{k=1}^{N} \frac{p_kU_k}{\sum_{k'=1}^N p_{k'} U_{k'}} F_k(w),
	\label{eq:Global_loss_distorted_reshaped}
\end{IEEEeqnarray}
which has the same solution as \eqref{eq:Global_loss_distorted}. Let us denote the solution to \eqref{eq:Global_loss_distorted_reshaped} by $\hat{w}^*$, and define $\alpha_k=p_k U_k$, $\alpha=\sum_{k=1}^N \alpha_k$, and $\alpha'_k = \frac{\alpha_k}{\alpha}$; therefore, $\sum_{k=1}^N \alpha'_k=1$. For brevity, we also define $H_{k,m}=\mathbf{1}\left( k\in\mathcal{S}(m), \text{SINR}_{k,m}>\theta \right)$, where we have ignored the time index of $\mathcal{S}$ since it is i.i.d. over different sampling steps.

Since we are using a different averaging approach, we must check \textbf{Lemma 2} and \textbf{Lemma 3}. However, \textbf{Lemma 4} is not affected by the averaging steps; thus, it still holds (after replacing $p_k$ with $\alpha'_k$). We also need to change $p_k$ to $\alpha'_k$ in definitions of $\bar{v}_t$, $\bar{w}_t$, ${g}_t$, and $\bar{g}_t$. It is worth reminding that, for sampling, we use {\em Scheme II} with $\hat{q}_k=p_k$.

Moreover, according to \eqref{eq:Compare-Averaging2}, when there is no successful transmission, the global model parameters at the BS do not change, which only affects the convergence rate (not the converging point). Therefore, to prove the convergence of \eqref{eq:Compare-Averaging2} to $\hat{w}^*$, we assume 
$\sum_{k'=1}^N \sum_{m'=1}^M H_{k',m'}>0$, i.e., at least one local update is available at the BS at each averaging step. Given $\sum_{k'=1}^N \sum_{m'=1}^M H_{k',m'}>0$,  we can write \eqref{eq:Compare-Averaging2} as
\begin{IEEEeqnarray}{rCl}
	w_{t} = \sum_{k=1}^{N}\sum_{m=1}^M 
	\frac{ H_{k,m} }
	{ \sum_{k'=1}^N \sum_{m'=1}^M H_{k',m'} } v_t^k, \qquad t\in\mathcal{I}_E.
	\nonumber 
	\label{eq:Compare-Averaging2-new-notation}
\end{IEEEeqnarray}

In the following, we provide a lemma that helps us derive \textbf{Lemma 2} and \textbf{Lemma 3} for the new averaging approach.
\begin{Lemma} For Scheme II sampling with $\left\{ \hat{q}_k=p_k \right\}$, we have
	\begin{IEEEeqnarray}{rCl}
		\mathbb{E}\left[ \sum_{m=1}^M \frac{ H_{k,m} }{ \sum_{k'=1}^N \sum_{m'=1}^M H_{k',m'} } \mid  \sum_{k'=1}^N \sum_{m'=1}^M H_{k',m'}>0 \right] = \alpha'_k. \nonumber 
	\end{IEEEeqnarray}
\end{Lemma}
\begin{IEEEproof}
		\begin{IEEEeqnarray}{rCl}
		\IEEEeqnarraymulticol{3}{l}{
		\mathbb{E}\left[ \sum_{m=1}^M \frac{ H_{k,m} }{ \sum_{k'=1}^N \sum_{m'=1}^M H_{k',m'} } \mid  \sum_{k'=1}^N \sum_{m'=1}^M H_{k',m'}>0 \right] } 
		\nonumber \\
		&=& 
		\sum_{m=1}^M \mathbb{E}_{H_{k,m}}\left[ \mathbb{E}\left[ \frac{ H_{k,m} }{ \sum_{k'=1}^N \sum_{m'=1}^M H_{k',m'} } \mid  \sum_{k'=1}^N \sum_{m'=1}^M H_{k',m'}>0, H_{k,m} \right] \right] 
		\nonumber \\
		&=& \sum_{m=1}^M \mathbb{P}\left( H_{k,m}=1 \mid  \sum_{k'=1}^N \sum_{m'=1}^M H_{k',m'}>0 \right) 
		\nonumber \\
		&& \qquad \times
		\mathbb{E}\left[  \frac{1}{ 1+\sum_{k'=1}^N \sum_{\substack{m'=1,\\ m'\neq m}}^M H_{k',m'} } \mid  \sum_{k'=1}^N \sum_{m'=1}^M H_{k',m'}>0, H_{k,m}=1 \right] \nonumber \\
		&=& \sum_{m=1}^M \sum_{i=0}^{M-1}\frac{1}{1+i} 
		\mathbb{P}\left( \sum_{k'=1}^N \sum_{\substack{m'=1,\\ m'\neq m}}^M H_{k',m'}=i, H_{k,m}=1 \mid \sum_{k'=1}^N \sum_{m'=1}^M H_{k',m'}>0\right).
		\label{eq:C1}
	\end{IEEEeqnarray}
	When $H_{k,m}=1$, we have $\sum_{k'=1}^N \sum_{m'=1}^M H_{k',m'}>0$. Moreover, resource allocation is i.i.d. over difference resource blocks, i.e. random variable $\sum_{k'=1}^N \sum_{\substack{m'=1,\\ m'\neq m}}^M H_{k',m'}$ is independent of random variable $H_{k,m}$.
	\begin{IEEEeqnarray}{rCl}
		\IEEEeqnarraymulticol{3}{l}{
		\mathbb{P}\left( \sum_{k'=1}^N \sum_{\substack{m'=1,\\ m'\neq m}}^M H_{k',m'}=i, H_{k,m}=1 \mid \sum_{k'=1}^N \sum_{m'=1}^M H_{k',m'}>0\right) } 
	    \nonumber  \\
		&=& \frac{ \mathbb{P}\left( H_{k,m}=1 \right) }  
		       { \mathbb{P}\left( \sum_{k'=1}^N \sum_{m'=1}^M H_{k',m'}>0 \right) } \mathbb{P}\left( \sum_{k'=1}^N \sum_{\substack{m'=1,\\ m'\neq m}}^M H_{k',m'}=i \right) 
		\nonumber \\       
		&\stackrel{\text{(a)}}{=}& \frac{ \alpha_k }
		       {1-(1-\alpha)^M} \binom{M-1}{i}\alpha^i(1-\alpha)^{M-1-i}, 
		\label{eq:C2}
	\end{IEEEeqnarray}
	where (a) is obtained using $\mathbb{P}\left( H_{k,m}=1 \right)=\alpha_k$ and $\mathbb{P}\left( \sum_{k=1}^N H_{k,m}=1 \right)=\alpha$. Finally, \textbf{Lemma 5} is obtained by substituting \eqref{eq:C2} in \eqref{eq:C1}.
\end{IEEEproof}

When $t\in\mathcal{I}_E$, we have $\bar{w}_t=w_t$; thus,
\begin{IEEEeqnarray}{rCl}
	\mathbb{E}\left[\bar{w}_t \mid \sum_{k'=1}^N \sum_{m'=1}^M H_{k',m'}>0 \right] &=&
	\mathbb{E}\left[ \sum_{k=1}^{N}\sum_{m=1}^M \frac{ H_{k,m} }{ \sum_{k'=1}^N \sum_{m'=1}^M H_{k',m'} } v_t^k \mid \sum_{k'=1}^N \sum_{m'=1}^M H_{k',m'}>0 \right] \nonumber \\
	&=& \sum_{k=1}^{N} \mathbb{E}\left[ \sum_{m=1}^M \frac{ H_{k,m} }{ \sum_{k'=1}^N \sum_{m'=1}^M H_{k',m'} } \mid \sum_{k'=1}^N \sum_{m'=1}^M H_{k',m'}>0 \right] v_t^k
	\nonumber \\
	&\stackrel{\text{(a)}}{=}& \sum_{k=1}^{N} \alpha'_k v_t^k = \bar{v}_t, 
	\label{eq:C3}
\end{IEEEeqnarray}
where the expectation is with respect to sampling and success event and (a) follows from \textbf{Lemma 5}. 

At $t+1\in\mathcal{I}_E$, we also have 
\allowdisplaybreaks{
\begin{IEEEeqnarray}{rCl}
	\IEEEeqnarraymulticol{3}{l}{ \mathbb{E}\left[ \left\| \bar{w}_{t+1} - \bar{v}_{t+1} \right\|^2 \mid \sum_{k'=1}^N \sum_{m'=1}^M H_{k',m'}>0 \right] }
	\nonumber \\
	&=& \mathbb{E}\left[ \left\| \sum_{k=1}^N 
	\frac{ \sum_{m=1}^M H_{k,m} }{ \sum_{k'=1}^N \sum_{m'=1}^M H_{k',m'} } \left( v_{t+1}^k - \bar{v}_{t+1} \right)  \right\|^2 \mid \sum_{k'=1}^N \sum_{m'=1}^M H_{k',m'}>0 \right]
	\nonumber \\
	&\le& \mathbb{E}\left[ \sum_{k=1}^N 
	\frac{ \sum_{m=1}^M H_{k,m} }{ \sum_{k'=1}^N \sum_{m'=1}^M H_{k',m'} } \left\| v_{t+1}^k - \bar{v}_{t+1} \right\|^2 \mid \sum_{k'=1}^N \sum_{m'=1}^M H_{k',m'}>0 \right]
	\nonumber \\
	&\stackrel{\text{(a)}}{=}& \mathbb{E}\left[ \sum_{k=1}^N \alpha'_k \left\| v_{t+1}^k - \bar{v}_{t+1} \right\|^2 \right] 
	= \mathbb{E}\left[ \sum_{k=1}^N \alpha'_k \left\| \left(v_{t+1}^k-w_{t+1-E}\right) - \left(\bar{v}_{t+1}-w_{t+1-E}\right) \right\|^2 \right]
	\nonumber \\
	&=& \mathbb{E}\left[ \sum_{k=1}^N \alpha'_k \left(\left\| v_{t+1}^k-w_{t+1-E} \right\|^2 + \left\| \bar{v}_{t+1}-w_{t+1-E} \right\|^2
	 -2 \left(v_{t+1}^k-w_{t+1-E}\right)^T \left(\bar{v}_{t+1}-w_{t+1-E}\right) \right) \right]
	\nonumber \\
	&=&  \mathbb{E}\left[ \sum_{k=1}^N \alpha'_k \left\| v_{t+1}^k-w_{t+1-E} \right\|^2 \right] - \mathbb{E}\left[ \left\| \bar{v}_{t+1}-w_{t+1-E} \right\|^2 \right]
	\nonumber \\
	&\le& \mathbb{E}\left[ \sum_{k=1}^N \alpha'_k \left\| v_{t+1}^k-w_{t+1-E} \right\|^2 \right] 
	= \mathbb{E}\left[ \sum_{k=1}^N \alpha'_k \left\| \sum_{i=t+1-E}^{t} \eta_i \nabla F_k (w_i^k,\xi_i^k) \right\|^2 \right] 
	\nonumber \\
	&\le& \mathbb{E}\left[ \sum_{k=1}^N \alpha'_k E \sum_{i=t+1-E}^{t} \left\| \eta_i \nabla F_k (w_i^k,\xi_i^k) \right\|^2 \right]
	\le  \sum_{k=1}^N \alpha'_k E^2 \eta_{t+1-E}^2 G^2 \le 4 \eta_{t}^2 E^2 G^2,
	\label{eq:C4}
\end{IEEEeqnarray}}
where, to drive (a), we use \textbf{Lemma 5}. The last line is also obtained similar to \eqref{eq:B2}.

From \eqref{eq:C3}, we understand that \textbf{Lemma 2} holds (for problem \eqref{eq:Global_loss_distorted_reshaped}). Also, from \eqref{eq:C4}, we understand that \textbf{Lemma 3} holds with $B=1$. As we discussed earlier, \textbf{Lemma 4} is also valid (after replacing $p_k$ with $\alpha'_k$). Thus, following the same procedure as in \textbf{Appendix B}, we can prove that with using \eqref{eq:Compare-Averaging2} at averaging steps, at time $T\in\mathcal{I}_E$, we have
\begin{IEEEeqnarray}{rCl}
	\mathbb{E}\left[ \left\| {w}_{T}-\hat{w}^* \right\|^2 \right] \le 
	\frac{ \max\left\{ 
		\frac{4}{\mu^2}\left( \sum_{k=1}^{N} \alpha_k^{\prime\,2}\sigma_k^2 + 6L\Gamma + 8(E-1)^2G^2 + 4E^2G^2 \right), 
		\gamma \left\| w_{0}-\hat{w}^* \right\|^2 \right\} }{\gamma+T}. \nonumber
\end{IEEEeqnarray}
Hence, the algorithm converges to the solution to \eqref{eq:Global_loss_distorted_reshaped}. 

\IEEEpeerreviewmaketitle
\bibliographystyle{IEEEtran}
\bibliography{IEEEabrv,Bibliography}

\end{document}